\documentclass{article}

\PassOptionsToPackage{numbers, compress}{natbib}

\usepackage[preprint]{neurips_2022}

\usepackage{microtype}
\usepackage{graphicx}
\usepackage{booktabs} % for professional tables

\usepackage{hyperref}
\usepackage{algorithm}

\usepackage{amsmath}
\usepackage{amssymb}
\usepackage{mathtools}
\usepackage{amsthm}
\usepackage{wrapfig}
\usepackage[capitalize,noabbrev]{cleveref}

\theoremstyle{plain}

\theoremstyle{definition}

\theoremstyle{remark}

\usepackage[textsize=tiny]{todonotes}

\usepackage[utf8]{inputenc} % allow utf-8 input
\usepackage{xcolor}
\usepackage{microtype}
\usepackage{graphicx}
\usepackage{slashbox}  
\usepackage{booktabs} % for professional tables (professional-quality tables)
\usepackage{algorithmic}
\usepackage{bm}
\usepackage{caption}
\usepackage{nicefrac}  % compact symbols for 1/2, etc.
\usepackage{multirow}
\usepackage{array}
\usepackage{dsfont}

\usepackage{times}
\usepackage{soul}
\usepackage{url}
\usepackage{newtxmath}
\let\OLDthebibliography\thebibliography
\renewcommand\thebibliography[1]{
  \OLDthebibliography{#1}
  \setlength{\parskip}{0pt}
  \setlength{\itemsep}{5pt plus 0.3ex}
}
\usepackage{subfigure}
\usepackage{colortbl}
\definecolor{mygraylite}{gray}{.94}
\definecolor{mygray}{gray}{.89}
\definecolor{darkergreen}{RGB}{21, 152, 56}
\definecolor{amber}{rgb}{1.0, 0.75, 0.0}
\definecolor{darkseagreen}{rgb}{0.56, 0.74, 0.56}
\definecolor{babyblueeyes}{rgb}{0.63, 0.79, 0.95}
\definecolor{burntsienna}{rgb}{0.91, 0.45, 0.32}
\usepackage{tabularx}
\usepackage{pifont}

\usepackage{pgfplots}
\usetikzlibrary{plotmarks}
\usepgfplotslibrary{fillbetween}
\pgfplotsset{compat=1.15}
\pgfplotsset{
    discard if not/.style 2 args={
        x filter/.code={
            \edef\tempa{\thisrow{#1}}
            \edef\tempb{#2}
            \ifx\tempa\tempb
            \else
                
            \fi
        }
    }
}
\pgfplotstableset{
    alias/MM-A/.initial=MM-A(alpha0.5),
    alias/MM-M/.initial=MM-M(alpha0.5),
    alias/UnMix-Mix-Code/.initial=UnMix+Mix-Code
}

\usepackage{enumitem} 
\newcommand{\hide}[1]{} %hide
\usepackage{xspace}
\newcommand*{\eg}{\emph{e.g.},\@\xspace}
\newcommand*{\ie}{\emph{i.e.},\@\xspace}
\newcommand*{\modelname}{{UnReMix}\@\xspace}
\newcommand*{\etc}{\emph{etc.}\@\xspace}

\newcommand\scalemath[2]{\scalebox{#1}{\mbox{\ensuremath{\displaystyle #2}}}}

\usepackage{cleveref}
\newcommand*{\romnum}[1]{\expandafter\@slowromancap\romannumeral #1@}
\newcommand{\rulesep}{\unskip\ \vrule\ }

\usepackage[T1]{fontenc}    % use 8-bit T1 fonts
\usepackage{amsfonts}       % blackboard math symbols
\usepackage[nomessages]{fp}
\usepackage{pgfkeys}
    \def\addlegendimage{\csname pgfplots@addlegendimage\endcsname}

\title{Hard Negative Sampling Strategies for Contrastive Representation Learning}

\author{
 Afrina Tabassum \\
  Department of Computer Science\\
  Virginia Tech\\
  \texttt{afrina@vt.edu} \\
   \And
   Muntasir Wahed \\
  Department of Computer Science\\
  Virginia Tech\\
  \texttt{mwahed@vt.edu} \\
  \And
  Hoda Eldardiry \\
  Department of Computer Science\\
  Virginia Tech\\
  \texttt{hdardiry@vt.edu} \\
  \And
  Ismini Lourentzou \\
  Department of Computer Science\\
  Virginia Tech\\
  \texttt{ilourentzou@vt.edu} \\
}

\begin{document}

\maketitle

\begin{abstract}
One of the challenges in contrastive learning is the selection of appropriate \textit{hard negative} examples, in the absence of label information. 
Random sampling or importance sampling methods based on feature similarity often lead to sub-optimal performance. 
In this work, we introduce \modelname, a hard negative sampling strategy that takes into account anchor similarity, model uncertainty and representativeness. 
Experimental results on several benchmarks show that \modelname improves negative sample selection, and subsequently downstream performance when compared to state-of-the-art contrastive learning methods. 
\end{abstract}

\section{Introduction}\label{sec:intro}
Due to the potential of alleviating data annotation costs and the substantial effort requirements in encoding domain-specific knowledge, self-supervised representation learning methods have attracted research effort, with recent contrastive learning frameworks even surpassing the downstream performance of supervised learning~\citep{chen2020simple, he2020momentum}.
Typically, contrastive methods aim at minimizing distances in feature space for similar example pairs (positive examples) and maximizing distances of dissimilar example pairs (negative examples) ~\citep{chopra2005learning}. 
There has been a growing interest in contrastive learning research, in particular for obtaining better data representations~\citep{chen2020simple, he2020momentum, robinson2020hard, zhu2020eqco, chuang2020debiased}.

Several recent studies investigate the impact of negative sampling strategies, with recent work showing that increasing the number of negative samples results in learning better representations. However, a few of the hardest negative samples tend to have the same label as the anchor, hampering the learning process \citep{he2020momentum,cai2020all}.
Hence, selecting appropriate informative hard negative examples is a crucial step for the success of contrastive learning. 

Various negative selection mechanisms have been proposed, that mainly aim to select discriminative negative examples based on the current learned feature representations, often used in conjunction with importance sampling or mixup interpolation~\citep{li2020prototypical,chen2020mocov2,chen2020simple,kalantidis2020hard,lee2020mix,kim2020mixco,shen2022unmix,robinson2020hard,xiong2020approximate,ma2021active}. 
Most contrastive methods either uniformly sample negatives, \ie assuming that all negative examples are equally important ~\citep{chen2020simple,he2020momentum,kalantidis2020hard}, compute importance scores based on feature similarity ~\citep{robinson2020hard,huynh2022fnc} or uncertainty~\citep{ma2021active}, or employ mixing of features in the feature/image space~\citep{lee2020mix,shen2022unmix,kim2020mixco,zhang2017mixup, ge2021robust,kalantidis2020hard,zhu2021improving}. As such, there is no clear notion of ``informativeness'' incorporated in the negative selection process. Particularly, prior works 
rarely consider the distance from the model decision boundary, \eg by incorporating model uncertainty, or representativeness of the selected negative examples, \eg whether selected negatives indeed represent the diverse distribution of negatives.

\begin{figure*}[t!]
    \centering
    \resizebox{.95\linewidth}{!}{
    \subfigure[Random]{
        \includegraphics[width=.33\linewidth]{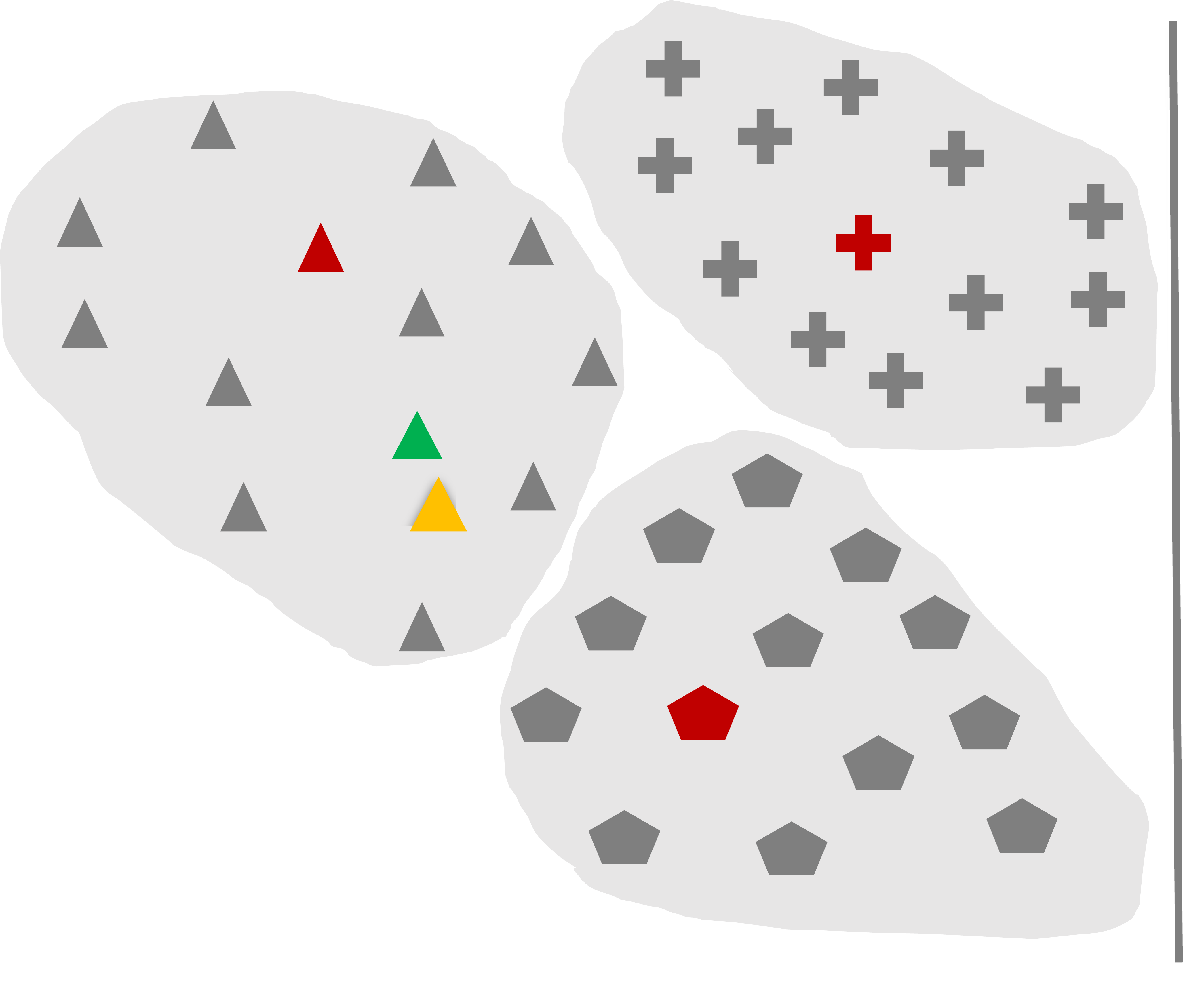}
        }
    \subfigure[Anchor Similarity]{\includegraphics[width=.33\linewidth]{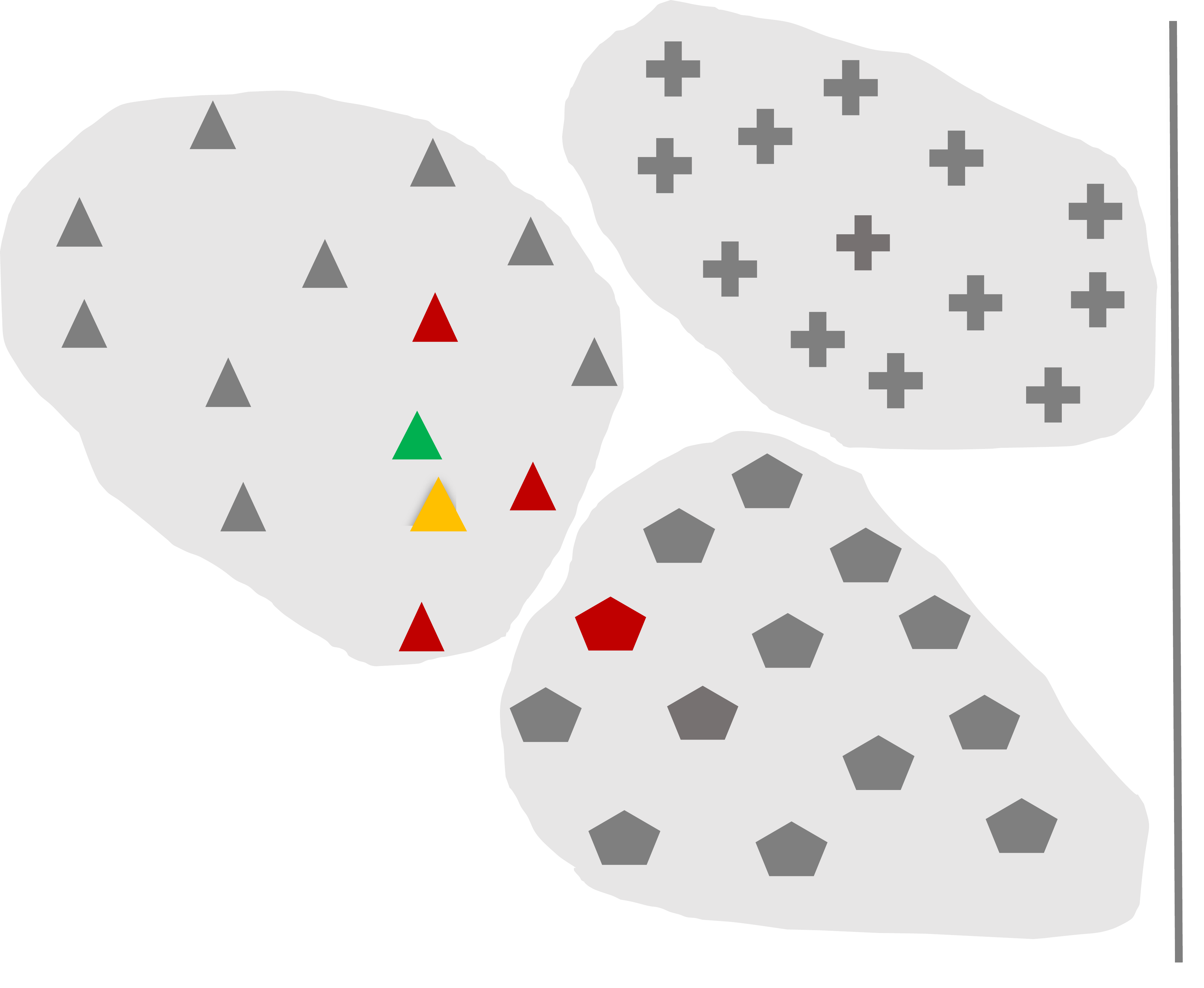}}
    \subfigure[\modelname (this work)]{\includegraphics[width=.33\linewidth]{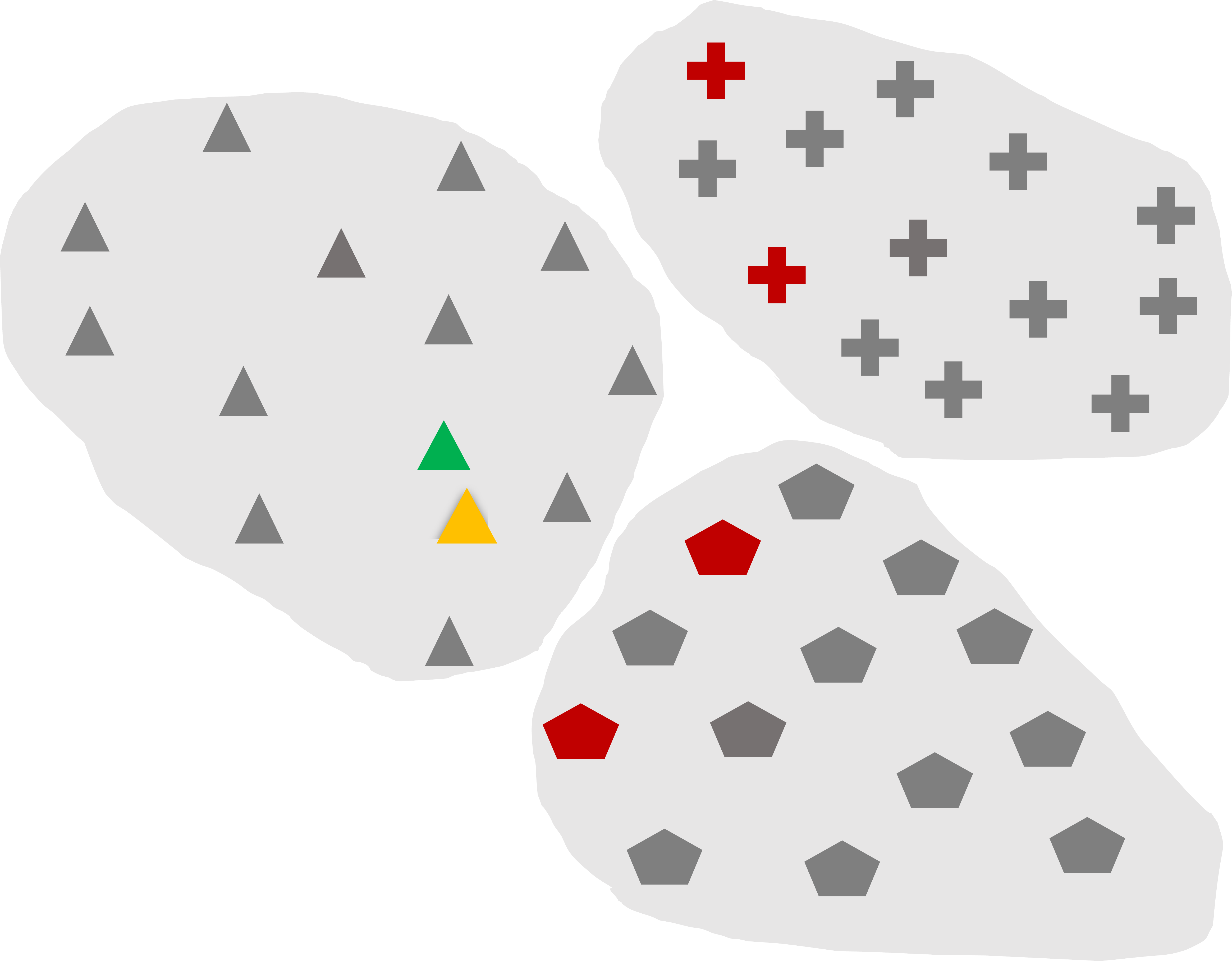}}}
    \vspace{-0.1in}
    \caption{Illustration of negative samples (\textcolor{red}{\textbf{red}} shapes) for an anchor (\textcolor{amber}{\textbf{yellow}} triangle) and its positive pair (\textcolor{green}{\textbf{green}} triangle), selected by three negative sampling techniques. Gray areas represent three clusters with different semantic labels. (a) Random sampling results in easy negatives being selected (1 \textcolor{red}{\textbf{triangle}}, 2 \textcolor{red}{\textbf{plus}}, 1 \textcolor{red}{\textbf{pentagon}}). (b) Methods that only consider anchor similarity may sample negatives that lie close to the anchor but also belong to the same semantic class (3 \textcolor{red}{\textbf{triangles}}, 1 \textcolor{red}{\textbf{pentagon}}). In contrast, (c) \modelname samples negatives that lie close to the decision boundary and are far from each other, \ie representing the whole data population (2 \textcolor{red}{\textbf{plus}}, 2 \textcolor{red}{\textbf{pentagons}}).}
    \vspace{-0.5cm}
    \label{fig:illustration_example}
\end{figure*}

Hard negative example selection in contrastive learning poses three challenges: (1) there is no label information available, hence there is a requirement for unsupervised strategies for instance selection, (2) an efficient sampling method should avoid false ``hardest'' negative samples, \ie samples that are most similar and originate from the same class as the anchor, and (3) an ideal set of negative examples should represent the whole population~\cite{robinson2020hard,cai2020all}. 
A selection mechanism should ideally capture all three properties: anchor similarity, model confidence and representativeness. Methods that only consider similarity with the anchor when sampling negative examples, \ie assuming that higher similarity aligns with higher importance, tend to select same-class negatives (Figure~\ref{fig:illustration_example}(b)), which can be detrimental to the representation learning \citep{cai2020all}. Besides, negative examples that lie closer to the decision boundary are naturally the hardest examples, encapsulating rich information about different class categories. On the other hand, representative negative examples help to learn global representations of the data distribution. Consequently, we argue that considering representativeness, along with uncertainty and anchor similarity, is helpful when selecting negative examples. 

To address the aforementioned limitations, this paper introduces \textbf{{Un}}certainty and \textbf{{Re}}presentativeness \textbf{{Mix}}ing (\textbf{\modelname}) for contrastive training, a method that combines importance scores that capture model uncertainty, representativeness, and anchor similarity. Specifically,  as illustrated in Figure~\ref{fig:illustration_example}(c), \modelname utilizes uncertainty to penalize false hard negatives and pairwise distance among negatives to select representative examples. To the best of our knowledge, we are the first to consider representativeness for hard negative sampling in contrastive learning in a computationally inexpensive way. We verify our method on several visual, text and graph benchmark datasets and perform comparisons over strong contrastive baselines.
Experimental and qualitative results demonstrate the effectiveness of our proposed approach. 

\textbf{Contributions:} The contributions of our work are summarized as follows:

\textbf{(1)} We delve into an empirical analysis of the efficacy of hard negative sampling strategies and feature-based importance sampling methods, observing that incorporating representativeness improves downstream performance.

\textbf{(2)} Based on our observations, we introduce an efficient method to calculate the representativeness of negative examples based on pairwise similarity and propose \modelname, a flexible and efficient hard negative sampling method for contrastive learning that selects hard informative negatives based on anchor similarity, model confidence and representativeness.

\textbf{(3)} We verify the effectiveness of the proposed method and show that \modelname improves downstream task performance and negative selection quality on several benchmarks in 3 domains (image, language
and graph). Qualitative analysis shows that the proposed method encourages sampling a diverse set of negatives, resulting in better performance.
\section{Related Work}\label{sec:related_work}
\paragraph{Contrastive Learning:} Recent work has largely contributed in developing contrastive self-supervised learning methods~\cite{chen2020simple,chuang2020debiased,chen2020mocov2,grill2020bootstrap, goyal2021self,zhuang2019local,caron2021emerging,li2021efficient,hjelm2018learning, oord2018representation,tian2020contrastive, Chen_2021_CVPR}, that have produced promising results in a variety of domains, from learning unsupervised cross-modal representations and video representations~\citep{ma2021active,Qian_2021_CVPR,Pan_2021_CVPR}, to natural language processing~\citep{Aberdam_2021_CVPR} and graph learning~\citep{you2020graph,qiu2020gcc}, and often achieving comparable results to supervised counterparts~\citep{chen2020simple, he2020momentum}.
\citet{chen2020simple} introduced SimCLR, a method that utilizes the examples of the current batch as negative samples. \citet{he2020momentum} and \citet{chen2020mocov2} introduced {MoCo}, a contrastive approach that incorporates a dynamic dictionary alongside with momentum update, and showed that larger dictionary sizes improve the accuracy on downstream tasks. \citet{he2020momentum} and its later improvements~\citep{chen2020mocov2, chen2021empirical} updated the dictionary by enqueueing the current batch and dequeuing the oldest batch. As a result, and according to \citet{ma2021active}, the dictionary will contain ``biased keys'', \ie keys of the same class. Moreover, \citet{saunshi2019theoretical} showed that increasing the dictionary size beyond a certain threshold may hamper the performance on downstream tasks. For audiovisual representations, \citet{ma2021active} proposed CM-ACC, where an actively sampled cross-modal dictionary look-up (to include keys of a wide range) is utilized. Cross-modal methods often cannot surpass strong baselines, \eg MoCo \citep{chen2020mocov2}, on unimodal datasets, because positive examples from a single modality (images) are less informative than cross-modal examples (video-audio pair), which can exchange auxiliary supervision signals between modalities.
In summary, research efforts can be largely divided into improvements over the contrastive loss calculation~\citep{chuang2020debiased,xie2021propagate,thota2021contrastive,oord2018representation,zhu2020eqco, Azabou2021, ko2021revisiting} and improvements over the hard positive/negative selection mechanisms ~\citep{shen2020mix,robinson2020hard,kalantidis2020hard,chen2020simple}. 
Our proposed \modelname method falls under the latter category and improves over feature-based contrastive importance sampling methods.
We compare \modelname with state-of-the-art hard negative sampling techniques, briefly described below.

\paragraph{Hard Negative Sampling:} Selection strategies for mining high-quality negative instances in contrastive learning have attracted substantial research interest, resulting in a wide range of contrastive methods proposed, \eg 
i-Mix~\citep{lee2020mix},
HCL~\citep{robinson2020hard}, Mochi~\citep{kalantidis2020hard},
AdCo~\citep{hu2021adco}, \etc In particular, AdCo~\citep{hu2021adco} maintains a separate global set for negative examples that is updated actively using the contrastive loss gradients with respect to each negative example. However, the set of negative examples remains the same for all the anchors. \citet{shah2021max} formulated the contrastive loss function as an SVM objective and utilized the support vectors as hard negatives, resorting to approximations to solve a computationally expensive quadratic equation for each anchor.  
\citet{cai2020all} performed an empirical study on the importance of negative samples in MoCo-V2 and found that only the hardest $5\%$ of the negatives are sufficient and necessary for achieving high accuracy and that same-class negatives can have detrimental effects on the representation learning process. \citet{robinson2020hard} proposed HCL, a method that calculates the importance of each negative example by considering feature similarity w.r.t. the anchor, and achieved improvements over MoCo-V2 and SimCLR.
\citet{kalantidis2020hard} synthesized negatives samples by mixing hard negatives (most similar negatives to the query) and achieved approximately $1\%$ improvement over MoCo-V2. Motivated by Mixup~\citep{zhang2017mixup}, a few methods create synthetic examples either by interpolating instances at an image/pixel or latent representation level \citep{kim2020mixco,shen2022unmix,zhu2021improving}, or by interpolating virtual labels~\citep{lee2020mix}.
\citet{ge2021robust} design negative examples using texture-based and patch-based non-semantic augmentation techniques. \citet{xiong2020approximate} propose an asynchronously-updated approximate nearest neighbor index for selecting negatives in text domains. 

HCL \citep{robinson2020hard} assigned more importance to negative examples that are similar to the query. Similarly, \citet{huynh2022fnc} relies on an anchor similarity threshold for filtering false negatives.
However, considering only feature similarity results in assigning more importance to the same-class negatives, \ie most likely false negatives which are detrimental to the representation learning process~\citep{cai2020all}. 
In contrast, \citet{ma2021active} selects negative examples with high model uncertainty. In active learning, \citet{ash2019deep} utilized the gradients of the loss function w.r.t. the model's most confident prediction as an approximation of uncertainty and theoretically showed that the gradient norm of the last layer of a neural network w.r.t. the predicted label provides a lower bound on gradient norms induced by any other label. \citet{ma2021active} utilized this measure to actively sample uncertain negatives when composing a memory bank. Our approach differs in that we leverage the gradients of the last layer as a model-based uncertainty measure so that our sampling method can assign more importance to the samples closer to the decision boundary, but also incorporate both anchor similarity and representativeness to capture other equally useful properties.

Overall, and to the best of our knowledge, none of the prior contrastive learning works jointly considers model confidence, anchor similarity and representativeness, let alone analyze the importance of interpolation among such components. To this end, we propose \modelname, a simple and efficient method that benefits from all aforementioned components when computing importance weights for negative examples.  Our experimental analysis shows that \modelname improves downstream performance and diversifies the negative example set. 

\section{Method}\label{sec:method}
\paragraph{Problem Formulation:}
Given an unlabeled dataset $\mathcal{X}$, we wish to learn an encoding function $f\colon \mathcal{X} \to \mathbb{R}^d$ that maps a data point $x_i\in \mathcal{X}$ to a $d$-dimensional embedding space, such that embeddings of similar instances $(x_i, x^{'}_{i})$ lie closer to each other, and vice versa. For a random subset (batch) of N positive pairs $\mathcal{X}_N \,{=}\, \{(\bar{x}_{i}, \tilde{x}_{i})\}_{i=1}^{N}$, where $\bar{x}_{i}, \tilde{x}_{i}$ are two augmented views of example $x_i$, the contrastive loss for learning the encoder $f$ is defined as  
\begin{align}
\scalemath{0.9}{
\mathcal{L}_{x_{i}} \,{=}\, -\log \frac{ \exp \big( s(\bar{x}_{i},\tilde{x}_{i}) / \tau \big) }
{ \exp \big( s(\bar{x}_{i},\tilde{x}_{i}) / \tau \big)
+ \sum\limits_{\tilde{x}_{j {\neq} i} \in \mathcal{X}_N} \exp \big( s(\bar{x}_{i},\tilde{x}_{j}) / \tau \big) },
}
\label{eq:nce}
\end{align}
where $s(x_{i}, {x}_{j}) \,{=}\, f(x_{i})^{\top} f({x}_{j}) / \lVert f(x_i) \rVert \lVert f(x_j) \rVert$ is the inner product of the normalized latent representations, and $\tau$ is a temperature scaling hyperparameter. Here, $\tilde{x}_{i}$ is referred as the positive sample for $\bar{x}_{i}$ and $\tilde{x}_{j {\neq} i} \in \mathcal{X}_N$ are the remaining instances, that are considered negative samples. 

The set of negative examples is typically selected by random sampling~\citep{chen2020simple, chen2020mocov2}. Recent works have individually proposed various ``hard'' negative mining or generation techniques, \eg based on perturbations in the input space~\citep{lee2020mix, shen2022unmix}, feature-based importance weights~\citep{robinson2020hard} or uncertainty-based sampling~\citep{ma2021active}. Yet, these methods consider only one selection indicator and hence achieve sub-optimal performance in learning contrastive representations. In this work, we propose a sampling technique, termed \modelname, that jointly considers both model-based uncertainty and representativeness to select negative examples. Below, we describe how \modelname captures the necessary properties for mining informative negative samples.  
\begin{figure}[t!]
\centering
  \includegraphics[width=\columnwidth]{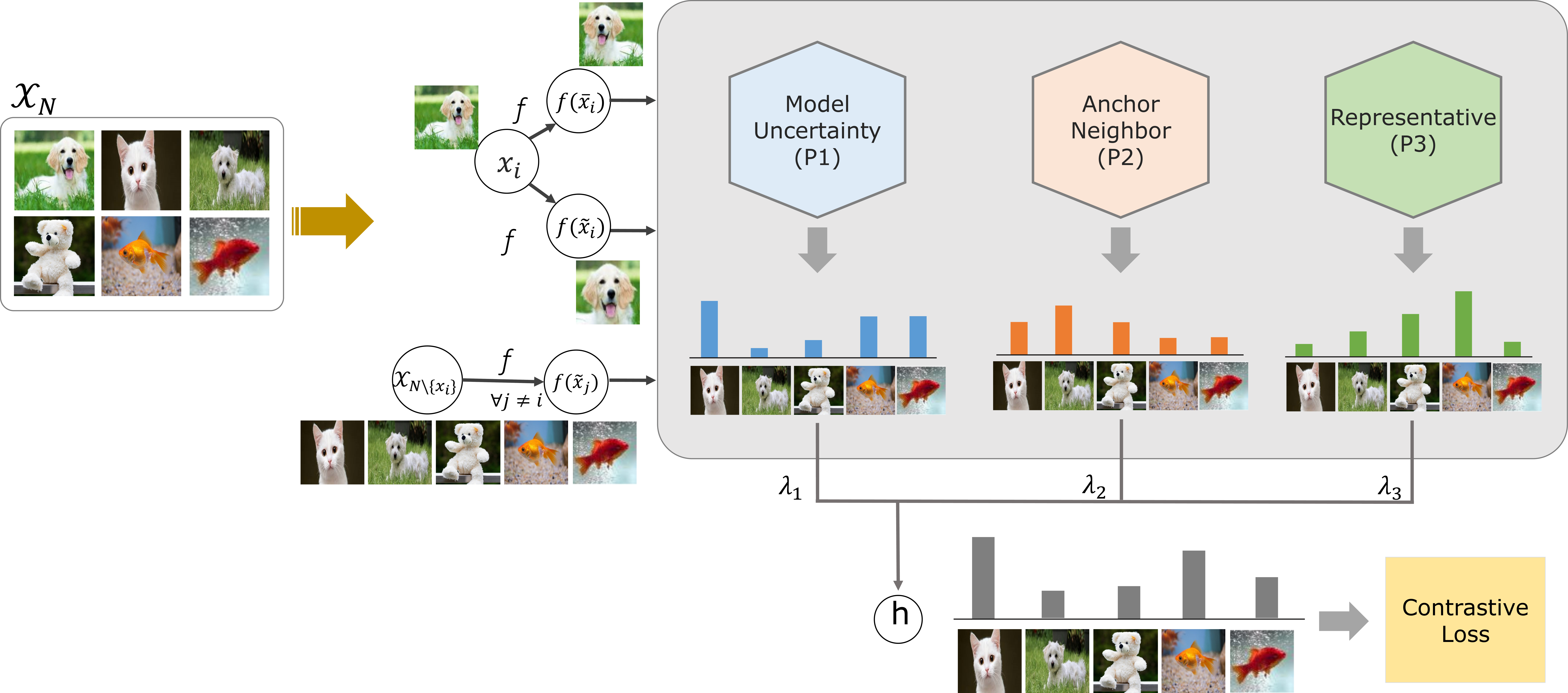}  
  \caption{Overview of the proposed hard negative sampling technique, \modelname. Given an anchor $x_i$ and a set of negative samples $\mathcal{X}_N \setminus x_i$, \modelname computes an importance score for each negative sample, by linearly interpolating between gradient-based uncertainty, anchor similarity and representativeness indicators, capturing desirable negative sample properties, \ie samples that are truly negative (P1), in close vicinity to the anchor (P2) and representative of the sample population (P3).}
\label{fig:model_diagram}
\vspace{-0.1in}
\end{figure}

\paragraph{\modelname Description:}
We wish to select high-quality informative hard negative examples that exhibit the following properties: 

\noindent \textbf{P1:} The ground-truth label of the selected negative example is different from the anchor label. We refer to P1 as the \textsc{TrueNegative} property.

\noindent \textbf{P2:} Hard negative examples resemble the anchor example, \ie the feature representations of the hardest negative examples lie close to the anchor in the embedding space. We refer to P2 as \textsc{anchor vicinity} property.

\noindent \textbf{P3:} Informative negative examples are representative of the sample population. In other words, semantically similar but not identical representative negative examples should be sufficient for contrastive training~\citep{cai2020all}. We refer to P3 as the \textsc{representativeness} property.

In summary, \modelname selects hard negative samples based on calculated importance scores. The higher the score is, the more informative the sample is assumed to be. The importance scores consist of three components: 
(1) a \textit{model-based component} that utilizes the loss gradients w.r.t. each negative sample as a measure of uncertainty and approximates \textbf{P1} by assigning more weight to the negative examples that lie closer to the decision boundary, 
(2) a \textit{feature-based component} that leverages the feature space geometry via instance similarity to select informative negative samples, that satisfy \textbf{P2}, and
(3) a \textit{density-based component} that assigns more weight to negatives examples that are more distant on average from other negative examples in the batch, and satisfies \textbf{P3}. We describe each component separately.

The lack of access to ground-truth label information makes it impossible to maintain \textbf{P1} (\textsc{TrueNegative}) completely. The challenge, therefore, lies in measuring the informativeness of negative samples without label information. 
Model uncertainty measures the degree of confidence of a model in its prediction \ie high model uncertainty corresponds to lower model confidence, and neural models typically assign higher uncertainty to examples closer to the decision boundary~\cite{liu2020simple}. We use this property to assign higher importance to negative examples, such that negative examples that are closer to the anchor but far from the decision boundary will have lower importance than negatives lying closer to the decision boundary.

Inspired by the use of similar information-theoretic metrics in metric learning~\citep{dutta2020unsupervised}, out-of-distribution detection~\citep{mundt2019open}, and reinforcement learning~\citep{zhao2019uncertainty}, we consider a gradient-based uncertainty metric. In particular, pseudo-labeling (as an implicit method for entropy minimization) and gradient-based uncertainty (where a smaller gradient norm corresponds to higher model confidence) are established in semi-supervised and
active learning \cite{lee2013pseudo,ash2019deep}. In addition, gradient-based uncertainty comes
with theoretical justifications for our chosen type of pseudo-labels
\cite{ash2019deep}.

More formally, we first define a pseudo-label space induced by the data distribution. We denote the most confident prediction for a negative example ${x}_{j {\neq} i} \in \mathcal{X}_N$ as its pseudo-label $\hat{y}_j$, and utilize the gradient of the last encoder layer w.r.t. this pseudo-label.
Specifically, we calculate the pseudo-posterior of the negative example $x_j$ via  
\begin{equation}\label{eq:posterior}
    p\left( {y}_j \left|~ \tilde{x}_j, \mathcal{X}_N \right. \right) \,{=}\, \frac{ \exp \big(s(\bar{x}_{i},\tilde{x}_{j})\big)}
    {\sum_{{\bar{x}_{i' \neq j} \in \mathcal{X}_N}} \exp \big(s(\bar{x}_{i'},\tilde{x}_{j})\big)},
\end{equation}
where $s(x_{i}, {x}_{j}) \,{=}\, f(x_{i})^{\top} f({x}_{j}) / \lVert f(x_i) \rVert \lVert f(x_j) \rVert$ is the inner product of the normalized representations of $x_{i}$ and $x_j$, respectively. Equation~(\ref{eq:posterior})
calculates the posterior as the similarity of a negative $x_j$ and all other examples $x_{i} \in \mathcal{X}_N$, considering them as individual anchors. 
We denote the most confident prediction as $\hat{y}_j \,{=}\, \arg\max_{k} \left[p\left( {y}_j \left| \tilde{x}_j, \mathcal{X}_N \right. \right)\right]_k$, where $\left[\cdot \right]_k$ corresponds to the class index, and calculate the gradient of the cross-entropy loss via
\begin{equation}\label{eq:gradient}
    g_{x_j}  \,{=}\, \frac{\partial}{\partial \theta_{last}} \left.\ell_{CE}\big(p\left( {y}_j \left| ~\tilde{x}_j, \mathcal{X}_N \right. \right), \hat{y}_j\big)\right|_{\theta=\theta_f},
\end{equation}
where $\ell_{CE}$ is the cross-entropy loss function and $\theta_{last}$ is the parameter vector of the last layer of encoder $f$. Intuitively, the gradient $g_{x_j}$ measures the model change caused by the negative example $x_j$. The more uncertain the model is about its prediction for a particular sample, the higher the update of the model parameters. Similarly, we compute $g_{x_i}$ for a specific anchor $x_i$. Finally, the uncertainty score of an example $x_j$ with respect to anchor $x_i$ is defined as $u(\tilde{x}_j,\bar{x}_i) \,{=}\, g_{x_i}^{\top} g_{x_j}$. 

To incorporate \textbf{P2} (\textsc{anchor vicinity}) in selecting hard negatives, we utilize instance similarity in the embedding space. Here, we use the inner product of the normalized vector representations as a similarity score for example $x_j$ with respect to anchor $x_i$, \ie $s(\tilde{x}_{j},\bar{x}_{i}) \,{=}\, f(\bar{x}_{i})^{\top} f(\tilde{x}_{j})/ \lVert f(\bar{x}_{i}) \rVert \lVert f(\tilde{x}_j) \rVert$.
This means that the more similar $x_j$ is to anchor $x_i$, the higher the importance of $x_j$ is~\citep{robinson2020hard}. Note that this contradicts \textbf{P1} by assigning more importance to negative examples with the same label as the anchor. Incorporating uncertainty via Equation~(\ref{eq:gradient}) alleviates this issue. Namely, by including the gradient vectors along with the feature representations in the calculation of the importance score for each negative sample, this specific uncertainty metric lessens the importance of the ``false'' negative samples. 
Even so, incorporating only uncertainty and anchor similarity does not allow for a negative sample set that is representative of the data population. The addition of an appropriate representativeness score aids in selecting informative hard representative negatives, maintaining \textbf{P3} (\textsc{representativeness}). Recent works utilize clustering of feature representations for selecting representative negative examples~\citep{li2020prototypical}. However, clustering after each update is computationally challenging and requires hyperparameters (number of clusters). To simplify the calculation of the representativeness score for a negative example, we instead compute its average distance from all other negative examples in the embedding space space. The representativeness score of an example $x_j$ given anchor $x_i$ is
\begin{equation}\label{eq:representative}
    r(\tilde{x}_{j},\bar{x}_{i}) \,{=}\, \frac{1}{N-2}
    \sum_{\substack{{j'}=1 \\ j' \notin \{i,j\}}}^{N}
    \Big(1 \,-\, s\left(\tilde{x}_j,\tilde{x}_{j'}\right) \Big).
    \vspace{-0.1cm}
\end{equation}
In our experiments, we observe that the proposed representativeness performs well and encourages sampling a diverse set of negatives.
Finally, we define the importance score of a negative example as follows:
\begin{align}
   w(\tilde{x}_{j},\bar{x}_{i}) \,{=}\, h\Big( u(\tilde{x}_{j},\bar{x}_{i}), s(\tilde{x}_{j},\bar{x}_{i}), r(\tilde{x}_{j},\bar{x}_{i})\Big).
\end{align}
Here $h$ is an aggregation function, \eg linear interpolation or attention weights. In our experiments, we use the latter and model the weight of each component as a learned hyper-parameter. Moreover, we present ablation studies for a variation with fixed equal weights.
\begin{algorithm}[!t]
\caption{\label{alg:calculate_importance} Pseudocode for \modelname}
\begin{algorithmic}
    \STATE \textbf{Input:} Dataset $\mathcal{X}$, batch size $N$, encoder $f$
    \FOR{batch $\mathcal{X}_N \,{=}\, \{(\bar{x}_{i}, \tilde{x}_{i})\}_{i=1}^{N} \sim\mathcal{X}$}
    \STATE $\texttt{loss} := 0$
    \FOR{$i,j = 1, \ldots, N$ and $j \neq i$}
        \STATE $~~$\textcolor{gray}{\# Acquire embeddings for positives and negatives, and calculate pairwise similarity scores}
        \STATE $~~$ $s(\tilde{x}_{j},\bar{x}_{i}) \,{=}\, f(\bar{x}_{i})^{\top} f(\tilde{x}_{j})/ \lVert f(\bar{x}_{i}) \rVert \lVert f(\tilde{x}_j) \rVert$
        \STATE $~~$$p\left( {y}_j \left|~ \tilde{x}_{j}, \mathcal{X}_N \right. \right) \,{=}\, \frac{ \exp \big(s(\bar{x}_{i},\tilde{x}_{j})\big)} {\sum_{{\bar{x}_{i' \neq j} \in \mathcal{X}_N}} \exp \big(s(\bar{x}_{i'},\tilde{x}_{j})\big)}$ ~\textcolor{gray}{\# Calculate pseudo-labels using Eq. (2)}
        \STATE $~~$$ g_{x_j}  \,{=}\, \nabla_{\theta_{last}} \left.\ell_{CE}\big(p\left( {y}_j \left| ~\tilde{x}_{j}, \mathcal{X}_N \right. \right), \hat{y}_j\big)\right|_{\theta=\theta_f}$ \textcolor{gray}{\# Calculate gradients using Eq. (3)} 
        \STATE $~~$$u(\tilde{x}_{j},\bar{x}_{i}) \,{=}\, g_{x_i}^{\top} g_{x_j}$ ~\textcolor{gray}{\# Calculate uncertainty score}
        \STATE $~~$ $r(\tilde{x}_{j},\bar{x}_{i}) \,{=}\, \frac{1}{N-2}
        \sum\limits_{\substack{{j'}=1 \\ j' \notin \{i,j\}}}^{N} \Big(1 - s(\tilde{x}_j,\tilde{x}_{j'})$ \Big)~\textcolor{gray}{\# Calculate representativeness score using Eq. (4)} 
        \STATE $~~$  $w(\tilde{x}_{j},\bar{x}_{i}) \,{=}\, \lambda_1 u(\tilde{x}_{j},\bar{x}_{i}) + \lambda_2 s(\tilde{x}_{j},\bar{x}_{i}) + \lambda_3 r(\tilde{x}_{j},\bar{x}_{i})$  ~\textcolor{gray}{\# Compute importance \small{($\lambda_1,\lambda_2,\lambda_3$ learnable)}}
         \STATE $~~$\textcolor{gray}{\# Compute contrastive loss with importance weights for negatives}
        \STATE $~~$\texttt{loss} += $\scalemath{0.7}{-\log \frac{ \exp \big( s(\bar{x}_{i},\tilde{x}_{i}) / \tau \big) }
{ \exp \big( s(\bar{x}_{i},\tilde{x}_{i}) / \tau \big)
+ \sum\limits_{\tilde{x}_{j {\neq} i} \in \mathcal{X}_N} w(\tilde{x}_{j},\bar{x}_{i}) \exp \big( s(\bar{x}_{i},\tilde{x}_{j}) / \tau \big) }}$
    \ENDFOR
    \STATE update the model to minimize the loss
    \ENDFOR
\end{algorithmic}
\end{algorithm}

Figure~\ref{fig:model_diagram} presents an overview of \modelname and Algorithm~\ref{alg:calculate_importance} provides the pseudocode for the calculation of importance scores.

\section{Experiments}\label{sec:experiments}

We evaluate \modelname on several benchmarks from various domains (image, graph and text data) and compare against state-of-the-art contrastive learning methods. The baseline set is the most representative w.r.t. hard negative selection or generation with competitive results over related work:

\noindent\textbf{MoCo}~\citep{chen2020mocov2}, more specifically MoCo-V2, a general dictionary-based contrastive learning method, for which negative examples are randomly sampled and stored in the dictionary. 

\noindent\textbf{Mochi}~\citep{kalantidis2020hard}, built on top of MoCo, this method generates synthetic hard negatives by mixing negatives stored in the dictionary. 

\noindent\textbf{SwAV}~\citep{caron2020unsupervised}, an online contrastive learning algorithm that uses a swapped prediction mechanism for clustering assignments. 

\noindent\textbf{i-Mix}~\citep{lee2020mix}, a regularization strategy that mixes data in both input and prediction level. 

\noindent\textbf{Patch-based NS}~\citep{ge2021robust}, a negative sample generation technique built upon MoCo-V2 that generates negative examples from the anchor using patch-based techniques.

\noindent\textbf{HCL}~\citep{robinson2020hard}, a hard negative selection strategy that improves negative selection  upon SimCLR~\citep{chen2020simple} by computing importance scores based on feature representations.  

\noindent\textbf{Un-Mix}~\citep{shen2022unmix}, a self-mixture strategy in the image space using Mixup~\citep{zhang2017mixup} and Cutmix~\citep{yun2019cutmix}. 

\subsection{Image Representations}
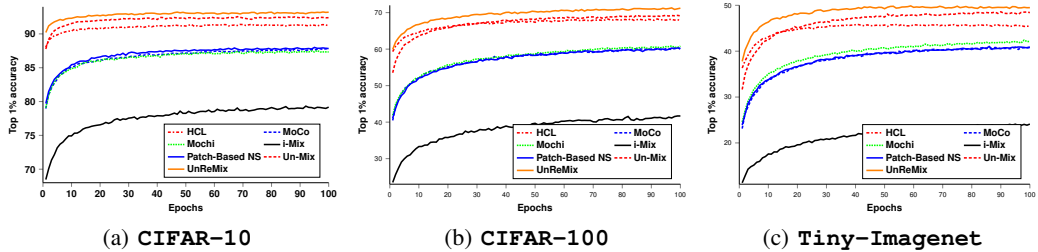
\begin{figure*}[t!]
    \centering
    \subfigure[\texttt{\textbf{CIFAR-10}}]{\resizebox{.32\linewidth}{!}{\begin{tikzpicture}[
        font=\bfseries\sffamily,
    ]
    \centering
    \begin{axis}[
        width=12cm,
        height=8cm,
        at={(0,0)},
        ymin=68,
        ymax=94,
        xmin=0,
        xmax=100,
        minor tick num =5,
        minor tick style={draw=none},
        minor grid style={thin,color=black!10},
        major grid style={thin,color=black!10},
        ylabel={Top 1\% accuracy},
        xlabel=Epochs,
        tick align=outside,
        axis x line*=middle,
        axis y line*=none,
        ylabel style={align=center},
        x tick label style={
            /pgf/number format/assume math mode},
        y tick label style={
        /pgf/number format/assume math mode},
        no markers,
        legend style={nodes=right},
        legend pos= south east,
        legend columns=2,
         every axis plot/.append style={ultra thick}
    ]
    \addplot [dashdotted, red] table [x=epoch, y=HCL, col sep=comma, mark=none] {data_files/LinearCifar10.csv};
    \addplot [densely dashed, blue] table [x=epoch, y=MoCo, col sep=comma, mark=none] {data_files/LinearCifar10.csv};
    \addplot [densely dotted, green] table [x=epoch, y=Mochi, col sep=comma, mark=none] {data_files/LinearCifar10.csv};
    \addplot table [x=epoch, y=IMix, col sep=comma, mark=none] {data_files/LinearCifar10.csv};
    \addplot table [x=epoch, y=Patch-Based NS, col sep=comma, mark=none] {data_files/LinearCifar10.csv};
    \addplot table [x=epoch, y=UnMix, col sep=comma, mark=none] {data_files/LinearCifar10.csv};
    \addplot [orange] table [x=epoch, y=UnMix-Mix-Code, col sep=comma, mark=none] {data_files/LinearCifar10.csv};
    \legend{HCL,MoCo, Mochi, i-Mix, Patch-Based NS, Un-Mix, \modelname}
    \end{axis}
\end{tikzpicture}} }
    \subfigure[\texttt{\textbf{CIFAR-100}}]{\resizebox{.32\linewidth}{!}{\begin{tikzpicture}[
        %Environment Cfg.
        font=\bfseries\sffamily,
    ]
    \begin{axis}[
        width=12cm,
        height=8cm,
        at={(0,0)},
        ymin=23,
        ymax=72,
        xmin=0,
        xmax=100,
        minor tick num =5,
        minor tick style={draw=none},
        minor grid style={thin,color=black!10},
        major grid style={thin,color=black!10},
        ylabel={Top 1\% accuracy},
        xlabel=Epochs,
        tick align=outside,
        axis x line*=middle,
        axis y line*=none,
        ylabel style={align=center},
        x tick label style={
            /pgf/number format/assume math mode, font=\sf\scriptsize},
        y tick label style={
        /pgf/number format/assume math mode, font=\sf\scriptsize},
        no markers,
        legend style={nodes=right},
        legend pos= south east,
        legend columns=2,
         every axis plot/.append style={ultra thick}
    ]
    \addplot [dashdotted, red] table [x=epoch, y=HCL, col sep=comma, mark=none] {data_files/LinearCifar100.csv};
    \addplot [densely dashed, blue] table [x=epoch, y=MoCo, col sep=comma, mark=none] {data_files/LinearCifar100.csv};
    \addplot [densely dotted, green] table [x=epoch, y=Mochi, col sep=comma, mark=none] {data_files/LinearCifar100.csv};
    \addplot table [x=epoch, y=IMix, col sep=comma, mark=none] {data_files/LinearCifar100.csv};
    \addplot table [x=epoch, y=Patch-Based NS, col sep=comma, mark=none] {data_files/LinearCifar100.csv};
    \addplot table [x=epoch, y=UnMix, col sep=comma, mark=none] {data_files/LinearCifar100.csv};
    \addplot [orange] table [x=epoch, y=UnMix-Mix-Code, col sep=comma, mark=none] {data_files/LinearCifar100.csv};
    \legend{HCL,MoCo, Mochi, i-Mix, Patch-Based NS, Un-Mix, \modelname}

    \end{axis}

    \end{tikzpicture}
    
%    }
} }
    \subfigure[\texttt{\textbf{Tiny-Imagenet}}]{\resizebox{.32\linewidth}{!}{\begin{tikzpicture}[
        %Environment Cfg.
        font=\bfseries\sffamily,
    ]
    \begin{axis}[
        width=12cm,
        height=8cm,
        at={(0,0)},
        ymin=11,
        ymax=50,
        xmin=0,
        xmax=100,
        minor tick num =5,
        minor tick style={draw=none},
        minor grid style={thin,color=black!10},
        major grid style={thin,color=black!10},
        ylabel={Top 1\% accuracy},
        xlabel=Epochs,
        tick align=outside,
        axis x line*=middle,
        axis y line*=none,
        ylabel style={align=center},
        x tick label style={
            /pgf/number format/assume math mode, font=\sf\scriptsize},
        y tick label style={
        /pgf/number format/assume math mode, font=\sf\scriptsize},
        no markers,
        legend style={nodes=right},
        legend pos= south east,
        legend columns=2,
         every axis plot/.append style={ultra thick}
    ]
    \addplot [dashdotted, red] table [x=epoch, y=HCL, col sep=comma, mark=none] {data_files/LinearTinyImagenet200k.csv};
    \addplot [densely dashed, blue] table [x=epoch, y=MoCo, col sep=comma, mark=none] {data_files/LinearTinyImagenet200k.csv};
    \addplot [densely dotted, green] table [x=epoch, y=Mochi, col sep=comma, mark=none] {data_files/LinearTinyImagenet200k.csv};
    \addplot table [x=epoch, y=IMix, col sep=comma, mark=none] {data_files/LinearTinyImagenet200k.csv};
    \addplot table [x=epoch, y=Patch-Based NS, col sep=comma, mark=none] {data_files/LinearTinyImagenet200k.csv};
    \addplot table [x=epoch, y=UnMix, col sep=comma, mark=none] {data_files/LinearTinyImagenet200k.csv};
    \addplot [orange] table [x=epoch, y=UnMix-Mix-Code, col sep=comma, mark=none] {data_files/LinearTinyImagenet200k.csv};
    \legend{HCL,MoCo, Mochi, i-Mix, Patch-Based NS, Un-Mix, \modelname}
    
    \end{axis}

    \end{tikzpicture}
    
 }}
    \vspace{-0.1in}
    \caption{Linear evaluation learning curves,
    \modelname (\textcolor{orange}{\textbf{orange}} lines) surpasses all baselines across all datasets.}
    \label{fig:linear_eval_over_epochs}
    \vspace{-0.3cm}
\end{figure*}

\paragraph{Linear Evaluation:} 
We follow prior contrastive learning works and train a linear classifier on frozen feature representations acquired from pre-trained contrastive models, and evaluate performance on the \texttt{CIFAR-10}, \texttt{CIFAR-100}, and \texttt{TINY-IMAGENET} datasets~\citep{cifar10_100,Le2015TinyIV}.
Figure~\ref{fig:linear_eval_over_epochs} depicts the consistent improvement of \modelname over all other baseline methods. 
Table~\ref{tab:acc_compare_with_std} presents the top-$1\%$ accuracy after fine-tuning the linear classifier for $100$ epochs. Results are averaged over multiple trials, \ie we report the mean and standard deviation over $10$ independent trials, and green arrows indicate relative gains over the next best method.
\modelname outperforms the best baseline (Un-Mix) by $0.76\%$ on \texttt{CIFAR-10}, $1.97\%$ on \texttt{CIFAR-100} and $1.01\%$ on \texttt{TINY-IMAGENET}. 
As far as the rest of the baselines, \modelname obtains on average an improvement of $2\%$ on \texttt{CIFAR-10}, $3.25\%$ on \texttt{CIFAR-100} and $3.83\%$ on \texttt{TINY-IMAGENET}.
\begin{table}[t!]
\newcommand*{\fct}[1]{\multicolumn{1}{>{\columncolor{white}\hspace*{-\tabcolsep}}c}{#1}}
\centering
\caption{Top 1\% accuracy comparison over baselines. Mean and standard deviation reported over 10 trials. Green arrows indicate relative gains over the next best method.}
\resizebox{\linewidth}{!}{
\begin{tabular}{l|l|l|l}
\hline
\selectfont \bf Method 
& \multicolumn{1}{c|}{\selectfont \bf \texttt{\textbf{CIFAR-10}}}
& \selectfont \bf $~~~$\texttt{\textbf{CIFAR-100}} 
& \selectfont \bf \texttt{\textbf{TINY-IMAGENET}} \\
\hline
SwAV~\citep{caron2020unsupervised}& \quad 76.90$\pm$0.02 	& \quad 43.60$\pm$0.01	&  \quad 29.00$\pm$0.10\\
i-Mix~\citep{lee2020mix} & \quad 79.26$\pm$0.18 	& \quad 41.58	$\pm$0.25&  \quad 24.10$\pm$0.02\\
MoCo~\citep{chen2020mocov2} & \quad 87.88$\pm$0.18	& \quad 59.96$\pm$0.20	&  \quad 40.76$\pm$0.40	\\
Patch-Based NS~\citep{ge2021robust} & \quad 87.86$\pm$0.06	& \quad 60.24$\pm$0.11	&  \quad 40.91$\pm$0.06	\\
Mochi~\citep{kalantidis2020hard} & \quad 87.33$\pm$0.12	& \quad 60.83$\pm$0.06	&  \quad 42.11$\pm$0.18\\
\rowcolor{mygraylite}  \cellcolor{white}HCL~\citep{robinson2020hard} & \quad 91.19$\pm$0.03& \quad 67.87$\pm$0.09	& \quad 45.62$\pm$0.07	\\
\rowcolor{mygraylite} \cellcolor{white}Un-Mix~\citep{shen2022unmix} & \quad 92.42$\pm$0.06	& \quad 69.15$\pm$0.06	& \quad 48.44$\pm$0.06\\
  \hline
 \rowcolor{mygray} \cellcolor{white}\bf\modelname & \quad \textbf{93.18$\pm$0.06}\color{darkergreen}$^{\bf\uparrow0.76}$	& \quad \textbf{71.12$\pm$0.09}\color{darkergreen}$^{\bf\uparrow1.97}$	& \quad \textbf{49.45$\pm$0.10}	\color{darkergreen}$^{\bf\uparrow1.01}$\\
\hline
\end{tabular}
}
\label{tab:acc_compare_with_std}
\end{table}

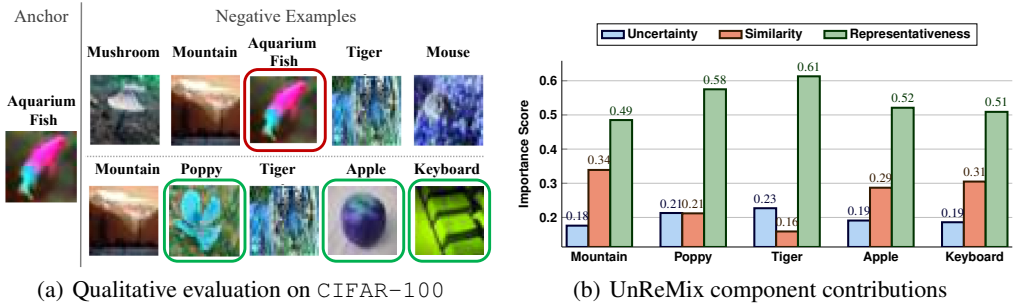
\begin{figure*}[t!]
    \centering
    \subfigure[Qualitative evaluation on \texttt{CIFAR-100}]{    \includegraphics[width=0.47\linewidth]{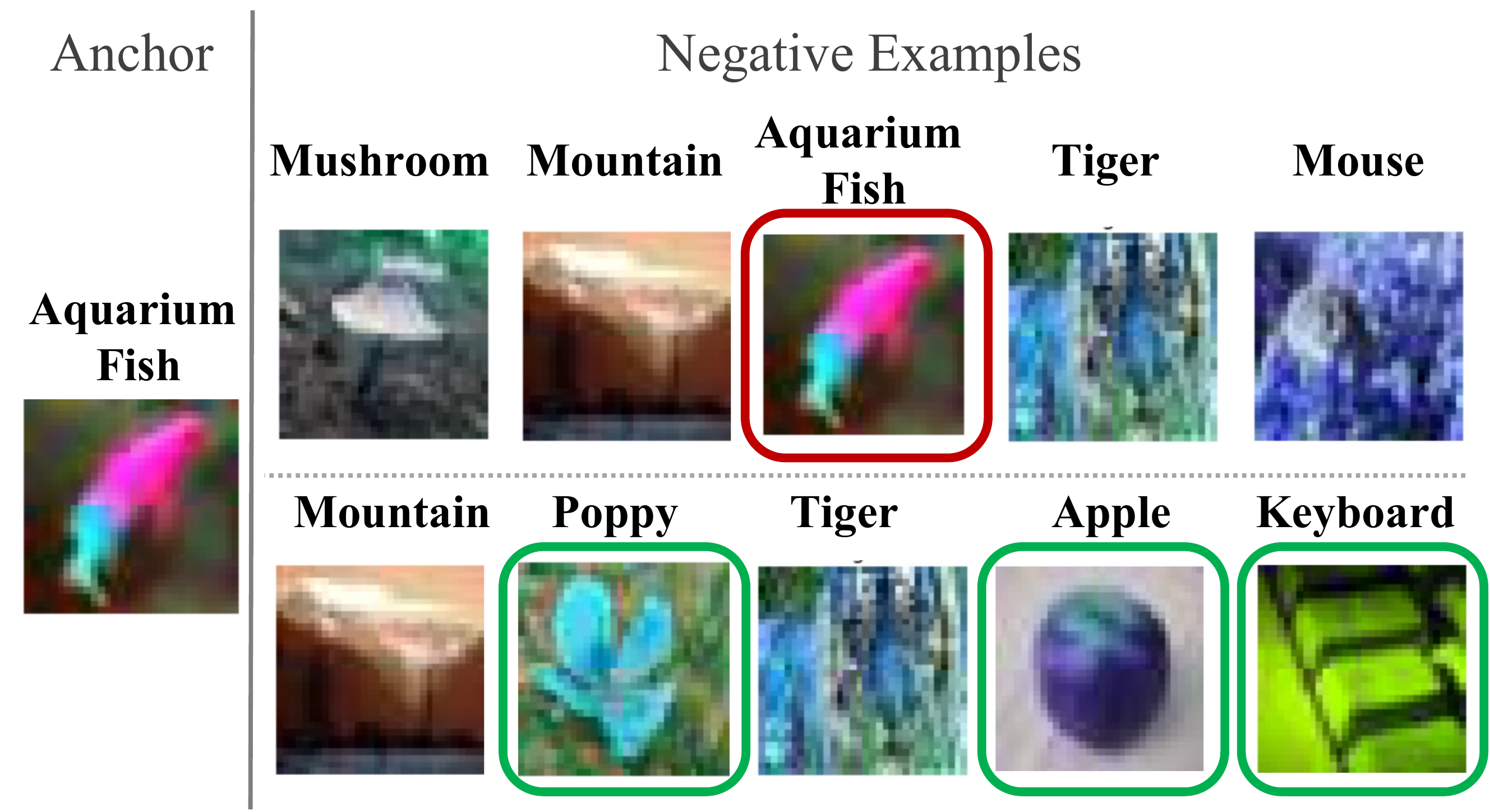}\label{fig:qualitative_sidebyside1}}
    \subfigure[\modelname component contributions]{
    \label{fig:qualitative_sidebyside2}
    \resizebox{0.48\linewidth}{!}{
        \begin{tikzpicture}[
        font=\bfseries\sffamily,
    ]
    \begin{axis}[
        ybar=0pt,
        width=\textwidth,
        height=0.48\textwidth,
        ymajorgrids, tick align=inside,
                symbolic x coords={Mountain,Poppy,Tiger,Apple,Keyboard},
        xtick=data,
        nodes near coords,
        ylabel={Importance Score},
        axis x line*=middle,
        axis y line*=none,
        ylabel style={align=center},
        y tick label style={
        /pgf/number format/assume math mode},
        bar width=0.6cm,
        legend style={
                        at={(0.5,1.2)},
                        anchor=north,
                        legend columns=-1,
                        /tikz/every even column/.append style={column sep=0.5cm}
                    },
        legend image code/.code={%
              \draw[#1] (0cm,-0.1cm) rectangle (0.6cm,0.1cm);
            } 
        legend columns=3,
        every axis plot/.append style={ultra thick}
    ]
    \addplot[blue!20!black,fill=babyblueeyes!80!white]  table[y=Uncertainty, col sep=comma]{data_files/normal_importance.csv};
    \addplot[orange!20!black,fill=burntsienna!80!white] table[y=Similarity, col sep=comma] {data_files/normal_importance.csv};
    \addplot[green!20!black,fill=darkseagreen!80!white] table[y=Representativeness, col sep=comma]{data_files/normal_importance.csv};
    \legend{Uncertainty, Similarity, Representativeness}
    \end{axis}
    \end{tikzpicture}
    }
    }
    \vspace{-0.1in}
    \caption{\textbf{(a)} Qualitative evaluation on \texttt{CIFAR-100}. Best viewed in color. HCL~\citep{robinson2020hard} (top row) has sampled the anchor as a negative example (\textcolor{red}{\textbf{red}} bounding box). \modelname (bottom row) samples more diverse true negative examples (\textcolor{green}{\textbf{green}} bounding boxes) and avoids false negative examples.
    \textbf{(b)} Individual contribution of each of the three importance score components, for the top-5 important negative examples selected by our method (bottom row of left figure).}
    \label{fig:qualitative_sidebyside}
\end{figure*}

\paragraph{Qualitative Analysis:} We compare the sampled negatives of both {HCL}~\citep{robinson2020hard} and \modelname.
Figure~\ref{fig:qualitative_sidebyside1} depicts the five most important negative examples sampled by {HCL} (top row) and \modelname (bottom row) for the same anchor example of class ``\textit{Acquarium Fish}''. As discussed earlier, \modelname samples more diverse negative examples and avoids false negatives. In contrast, {HCL} has sampled the anchor as negative (indicated with a red bounding box). As mentioned, \modelname components (uncertainty, feature similarity, and representativeness) are aggregated with learned hyper-parameters as weights. Figure~\ref{fig:qualitative_sidebyside2} depicts the contribution of each component in calculating the overall importance score of the top five important negative examples selected by \modelname. We can observe that the representativeness component contributes the most. Additional qualitative examples can be found in the Appendix.

\begin{table}[t]
\newcommand*{\fct}[1]{\multicolumn{1}{>{\columncolor{white}\hspace*{-\tabcolsep}}c}{#1}}
\centering
\caption{Top-$1\%$ accuracy of different \modelname variations.}
\resizebox{0.8\textwidth}{!}{
\begin{tabular}{l|l|l|l}
\hline
\fontsize{9pt}{1em}\selectfont \bf Components of ${w}$
& \multicolumn{1}{c|}{\fontsize{8pt}{1em}\selectfont \bf \texttt{\textbf{CIFAR-10}}}
& \fontsize{8pt}{1em}\selectfont \bf \texttt{\textbf{CIFAR-100}}
& \fontsize{8pt}{1em}\selectfont \bf \texttt{\textbf{TINY-IMAGENET}} \\
\hline 
Feature similarity (HCL) & \quad 91.19	& \quad 67.86	&  \quad 45.61	\\
Uncertainty & \quad 91.88	& \quad 68.77	&  \quad 47.44	\\
Representativeness & \quad 92.02 	& \quad 68.60	&  \quad 48.03\\
 \hline
\rowcolor{mygray} \cellcolor{white}  All (\modelname) & \quad \textbf{91.99} & \quad \textbf{69.43} & \quad \textbf{48.12}\\
\hline
\end{tabular}
}
\label{tab:individiual_component}
\vspace{-0.2cm}
\end{table}

\begin{table}[t]
\newcommand*{\fct}[1]{\multicolumn{1}{>{\columncolor{white}\hspace*{-\tabcolsep}}c}{#1}}
\centering
\caption{Top-1\% accuracy comparison over \modelname variations with learned or fixed equal weights for each component.}
\resizebox{0.8\textwidth}{!}{
\begin{tabular}{l|l|l|l}
\hline
\fontsize{9pt}{9pt}\selectfont \bf Method 
& \multicolumn{1}{c|}{\fontsize{9pt}{9pt}\selectfont \bf \textbf{\texttt{CIFAR-10}}}
& \fontsize{9pt}{9pt}\selectfont \bf \textbf{\texttt{CIFAR-100}} 
& \fontsize{9pt}{9pt}\selectfont \bf \textbf{\texttt{TINY-IMAGENET}} \\
\hline
\modelname$_{Fixed}$ & \quad 91.95$\pm$0.06	& \quad \textbf{69.27$\pm$0.13}	&  \quad 47.29$\pm$0.14	\\
\modelname$_{Learned}$ & \quad \textbf{92.00$\pm$0.06}	& \quad 68.78$\pm$0.15	&  \quad \textbf{47.85$\pm$0.09}	\\
\hline
\end{tabular}
}
\label{tab:acc_compare_aggregation_function}
\vspace{-0.1cm}
\end{table}

\paragraph{Importance score components:}
We perform an ablation analysis for each of the score components, \ie uncertainty (${u}$), similarity (${s}$), and representativeness (${r}$). We train different variations of \modelname, with one component at a time. Table~\ref{tab:individiual_component} presents the top-1\% accuracy of linear evaluation on the \texttt{CIFAR-10}, \texttt{CIFAR-100} and \texttt{TINY-IMAGENET} datasets, respectively. Note that including only feature similarity is essentially similar to HCL~\citep{robinson2020hard}. We notice that including uncertainty and representativeness separately outperform HCL. Moreover, adding all components (\modelname) results in performance improvements across all datasets.

\paragraph{Aggregation Function:}
\modelname combines uncertainty, similarity and representativeness in a linearly interpolated importance score, computed for each example via attention weights that are learned during training. Table~\ref{tab:acc_compare_aggregation_function} presents the accuracy of linear evaluation for \modelname and a variation with fixed equal weights. We observe that fixed equal weights perform comparatively well. We also present the evolution of weights for each  \modelname component in the supplementary material (Appendix ~\ref{sup:weights_over_epochs}). We leave further analysis with more variations to future work.
\begin{table}[!t]
\caption{Classification accuracy on CR, SUBJ, MPQA, TREC, MSRP downstream tasks and test Pearson Correlation for Semantic-Relatedness (SICK) task.  Sentence representations are learned using Quick-Thought (QT) vectors on the BookCorpus dataset and evaluated on six classification tasks.  Evaluation with 10-fold cross-validation for binary classification tasks (CR, SUBJ, MPQA) and over multiple trials for the remaining tasks (TREC, MSRP).
}
\resizebox{\textwidth}{!}{
\begin{tabular}{l|llllllll}
\hline
 \fontsize{9pt}{9pt}\selectfont\textbf{Method} &  \fontsize{9pt}{9pt}\selectfont\textbf{SICK} & \fontsize{9pt}{9pt}\selectfont\textbf{CR} & \fontsize{9pt}{9pt}\selectfont\textbf{SUBJ} & \fontsize{9pt}{9pt}\selectfont\textbf{MPQA} & \fontsize{9pt}{9pt}\selectfont\textbf{TREC} & \multicolumn{2}{c}{\fontsize{9pt}{9pt}\selectfont\textbf{MSRP}}
 \\
& & & & & &  \fontsize{9pt}{9pt}\selectfont\textbf{(Acc)} & \fontsize{9pt}{9pt}\selectfont\textbf{(F1)} \\
\hline
\hline
\fontsize{9pt}{9pt}\selectfont QT & \fontsize{9pt}{9pt}\selectfont 67.7 & \fontsize{9pt}{9pt}\selectfont 67.5 & \fontsize{9pt}{9pt}\selectfont 79.9 & \fontsize{9pt}{9pt}\selectfont\textbf{80.3} & \fontsize{9pt}{9pt}\selectfont 66.0 & \fontsize{9pt}{9pt}\selectfont 68.0 & \fontsize{9pt}{9pt}\selectfont 80.1
\\
\fontsize{9pt}{9pt}\selectfont HCL & \fontsize{9pt}{9pt}\selectfont 60.6 & \fontsize{9pt}{9pt}\selectfont 62.7 & \fontsize{9pt}{9pt}\selectfont 74.1 &\fontsize{9pt}{9pt}\selectfont 79.3 &\fontsize{9pt}{9pt}\selectfont 58.6 &\fontsize{9pt}{9pt}\selectfont 68.3 &\fontsize{9pt}{9pt}\selectfont 79.8
\\
\hline

\rowcolor{mygray}\cellcolor{white}\fontsize{9pt}{9pt}\selectfont\modelname  & \fontsize{9pt}{9pt}\selectfont\textbf{74.7}\color{darkergreen}$^{\bf\uparrow6.97}$ & 
\fontsize{9pt}{9pt}\selectfont \textbf{78.0}\color{darkergreen}$^{\bf\uparrow10.5}$ & \fontsize{9pt}{9pt}\selectfont \textbf{86.8}\color{darkergreen}$^{\bf\uparrow6.9}$ & \fontsize{9pt}{9pt}\selectfont 78.7 & \fontsize{9pt}{9pt}\selectfont \textbf{82.8}\color{darkergreen}$^{\bf\uparrow16.8}$ & \fontsize{9pt}{9pt}\selectfont \textbf{70.7}\color{darkergreen}$^{\bf\uparrow2.7}$ & \fontsize{9pt}{9pt}\selectfont \textbf{80.9}\color{darkergreen}$^{\bf\uparrow0.76}$
\\
\hline
\end{tabular}
}
\label{tab:sent_comps}
\end{table}

\subsection{Sentence Representations}
We evaluate our model on learning sentence representations using the Quick-Thought (QT) vectors~\citep{logeswaran2018an}, following the same experimental setting as ~\citet{logeswaran2018an}. Specifically, we train sentence embeddings using the BookCorpus dataset~\citep{kiros2015skip} and evaluate the learned embeddings on six downstream tasks: semantic relatedness (SICK), product reviews (CR), subjectivity classification (SUBJ), opinion polarity (MPQA), question type classification (TREC), and paraphrase identification (MSRP). Results are reported in Table~\ref{tab:sent_comps}, with \modelname outperforming baselines in all tasks. 

\subsection{Graph Representations}
We also evaluate \modelname on a graph representation task. Using the experimental settings of HCL~\citep{robinson2020hard}, we utilize the InfoGraph method~\citep{sun2019infograph} as the baseline and fine-tune an SVM readout function on the learned embeddings using $\beta=1$ (HCL hyperparameter) for six graph datasets.  Overall InfoGraph performs better in 3 out of the 6 benchmarks, and HCL has the best performance 2 out of 6 times. Our method works better or is comparable to InfoGraph and HCL in 2 benchmark datasets. We hypothesize that there is less variability in some of the graph datasets and thus the representative component is not utilized completely.
\begin{figure}[t!]
\centering
  \includegraphics[width=\textwidth]{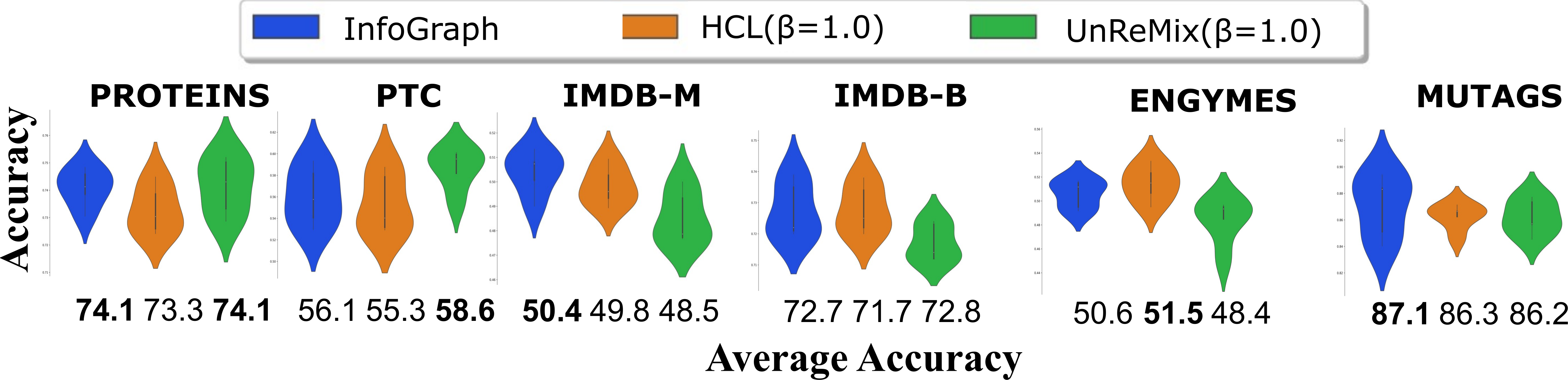}
  \vspace{-0.4cm}
  \caption{Classification Accuracy for six benchmark datasets. Results reported are averaged over 5 independent runs, each run with 10-fold cross-validation. Methods with the best performance are highlighted with bold.}
\label{fig:graph_results}
\end{figure}

\section{Conclusion}\label{sec:conlusion}
In this work, we present \modelname, a hard negative selection strategy that samples informative negative examples for contrastive training. We define the notion of ``informativeness'' by utilizing feature representations, model uncertainty and representativeness. As a measure of uncertainty, we extract the gradients of the loss function w.r.t. a computed pseudo-posterior for the negative examples. In addition, we utilize the average distance of each negative example from all other examples as a measure of representativeness. Feature representations from the last encoder layer are utilized in computing anchor similarity. Our approach interpolates these three indicators to determine the importance of negative samples. 

Through experimental analysis on a variety of visual, sentence and graph downstream benchmarks, we showcase that our proposed approach, \modelname, outperforms previous state-of-the-art contrastive learning methods, that either rely on random sampling for selecting negative samples, or on importance sampling calculated solely via feature similarity. In the future, we hope to evaluate our method in multi-modal large-scale benchmark datasets and extend \modelname to prototypical and graph contrastive learning.

\bibliography{biblio}
\bibliographystyle{plainnat}

\newpage
\appendix
\section{Training and hyper-parameter details}\label{sup:training_details}
\paragraph{Image representation:}For all models, we adopt the training setup of HCL~\citep{robinson2020hard}. More specifically, we use Resnet-50~\citep{he2016deep} as the base encoder, followed by a projection head or reducing the dimensionality of the feature representations from 2048 to 128. We train with Adam~\citep{kingma2014adam}, $10^{-3}$ learning rate and  $10^{-6}$ weight decay. While our method is implemented on top of Un-Mix, the underlying importance score computations are fairly general and can be easily incorporated into any contrastive learning method. 
We pre-train all models for 400 epochs with a batch size of 256.  All experiments are performed on an {NVIDIA T4} GPU with 16GB memory.

\paragraph{Sentence representation:}Similar to ~\citet{robinson2020hard}, we built upon the official experimental settings of quick-thoughts vectors (\url{https://github.com/lajanugen/S2V}). For each anchor sentence, the previous and next $k$ sentences (hyper-parameter) are considered positive examples, and all other examples are considered negative examples. Subsequently, Equation~\ref{eq:nce} is used as loss function to learn the model with Adam optimizer~\citep{kingma2014adam}, batch size of $400$, learning rate of $5 \times 10^{-4}$, and sequence length of $30$. We use the default values for all other hyper-parameters and implement our importance calculation in \texttt{s2v-model.py}. Since the official BookCorpus dataset~\cite{kiros2015skip} is not available, we use an unofficial version obtained from \url{https://github.com/soskek/bookcorpus}, and following  instructions from \citet{robinson2020hard}.

\paragraph{Graph representation:} We adopt the code of ~\citet{robinson2020hard} (\url{https://github.com/joshr17/HCL/tree/main/graph}) and incorporate our importance score calculation mechanism in \texttt{gan$\_$losses.py}. We utilize all datasets downloaded from \url{www.graphlearning.io}. For a fair comparison to the original \texttt{InfoGraph} method and HCL~\cite{robinson2020hard}, we train all the models using the same hyper-parameters values. Following \citet{robinson2020hard} we use the GIN architecture~\cite{xu2018powerful} with $K= 3$ layers and embedding dimension $d= 32$, trained for 200 epochs with  $128$ batch size, Adam optimizer, $10^{-3}$ learning rate, and $10^{-6}$ weight decay. Experiments in Figure~\ref{fig:graph_results} are reported over 10 experimental trials.  
\section{Ablation Studies of Image Representations}
\begin{wrapfigure}{r}{0.5\columnwidth}
\centering
\resizebox{.3\textwidth}{!}{
\begin{tikzpicture}
\begin{axis}[
  xlabel=\# of Classes,
  ylabel=KNN Accuracy]
\addplot table [x=k, y=$KNN_Acc$]{data_files/k_cluster_table.dat};
\end{axis}
\end{tikzpicture}
}
\caption{Top-5\% KNN Accuracy when training with negative examples sampled from $k=\{10,20,30,40,50\}$ classes ($x$-axis).}
\label{fig:knn_acc_over_k}
\end{wrapfigure}
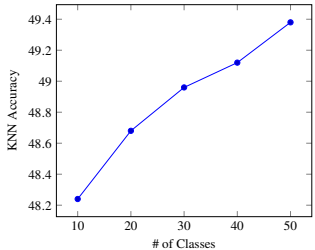\subsection{Importance of representativeness}\label{sup:accuracy_over_classes}
To demonstrate the importance of selecting representative negative examples, we experiment with the diversity of the negative samples. We trained a SimCLR~\cite{chen2020simple} model with all default settings and use a modified sampling process. More specifically, we sample negative examples of classes from $0$ to $k$ classes, where $k$ is the hyper-parameter for controlling diversity of negative examples and $k=\{10,20,30,40,50\}$. Figure~\ref{fig:knn_acc_over_k} presents the KNN Accuracy of SimCLR trained for $100$ epochs. We observe that as $k$ increases, \ie the diversity of selected negative examples increases, accuracy also increases. Hence, representativeness \ie diversity of the negative examples contributes to learning better representations. 

\subsection{Weight variation for different datasets}\label{sup:weights_over_epochs}
\begin{table}[t]
\resizebox{\linewidth}{!}{
\centering
\begin{tabular}{cccc}
& \textbf{\huge{Similarity}} & \textbf{\huge{Uncertainty}}  & \textbf{\huge{Representativeness}} \\ 
\rotatebox[origin=l]{90}{\huge{\textbf{\texttt{CIFAR-10}}}} &\includegraphics[width=0.6\textwidth,height=0.4\textwidth]{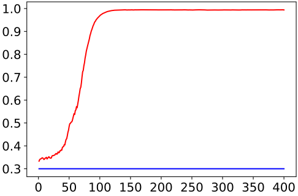} & \includegraphics[width=0.6\textwidth,height=0.4\textwidth]{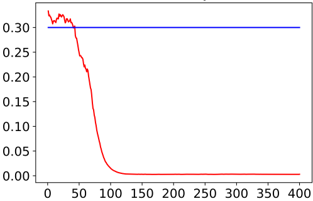} & \includegraphics[width=0.6\textwidth,height=0.4\textwidth]{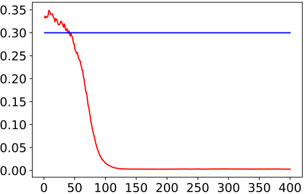}\\
\rotatebox[origin=l]{90}{\huge{\textbf{\texttt{CIFAR-100}}}} & \includegraphics[width=0.6\textwidth,height=0.4\textwidth]{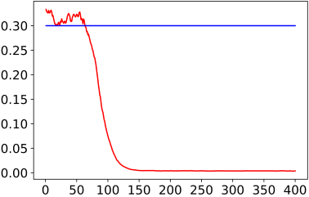} & \includegraphics[width=0.6\textwidth,height=0.4\textwidth]{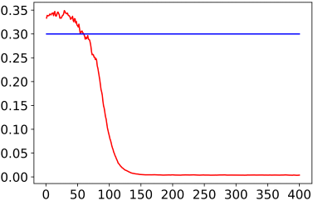} & \includegraphics[width=0.6\textwidth,height=0.4\textwidth]{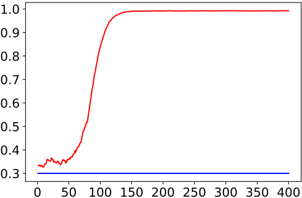}\\
\rotatebox[origin=l]{90}{\huge{\textbf{\texttt{TINY-IMAGENET}}}} & \includegraphics[width=0.6\textwidth,height=0.4\textwidth]{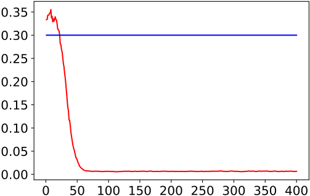} & \includegraphics[width=0.6\textwidth,height=0.4\textwidth]{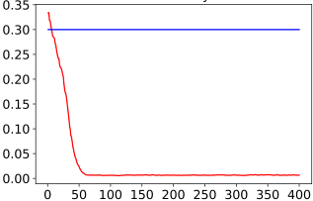} & \includegraphics[width=0.6\textwidth,height=0.4\textwidth]{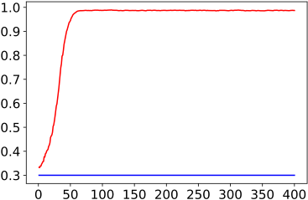}
\end{tabular}
}
\vspace{0.1cm}
\captionof{figure}{Weight (y-axis) of each component over 400 epochs (x-axis) for  \modelname (\textcolor{red}{\textbf{red}} lines) and a variant with fixed equal weights, \modelname$_{Fixed}$ (\textcolor{blue}{\textbf{blue}} lines). Results on \texttt{CIFAR-10}, \texttt{CIFAR-100} and \texttt{TINY-IMAGENET}, best viewed in color.} 
\label{fig:weight_over_epochs}
\end{table}
\modelname combines uncertainty, similarity and representativeness in one importance score for each example via attention weights that are learned during training. Figure~\ref{fig:weight_over_epochs} presents the weights for each component as learning progresses. For both \texttt{CIFAR-100} and \texttt{TINY-IMAGENET} weights converge after 100 epochs, assigning most importance to representativeness. However, for \texttt{CIFAR-10}, weights for similarity are higher than the weights for representativeness. In addition, uncertainty weights have converged to lower values across all three datasets. We could infer that the weights for each component would vary depending on the dataset and by learning the weights (with an attention mechanism) \modelname can converge to the optimal combination. Finally, we observe that the weights of the attention mechanism converge quickly to a local minima solution because of the softmax function.  Removing softmax makes convergence slower but does not affect accuracy.

\begin{table}[h]
\newcommand*{\fct}[1]{\multicolumn{1}{>{\columncolor{white}\hspace*{-\tabcolsep}}c}{#1}}
\centering
\caption{Top-1\% accuracy of \modelname variations with gradients computed with (a) Cross-Entropy loss (CE) and (b) NT-Xent loss.}
\resizebox{0.8\textwidth}{!}{
\begin{tabular}{l|l|l|l}
\hline
\fontsize{8pt}{1em}\selectfont \bf Method 
& \multicolumn{1}{c|}{\fontsize{8pt}{1em}\selectfont \bf \textbf{\texttt{CIFAR-10}}}
& \fontsize{8pt}{1em}\selectfont \bf \textbf{\texttt{CIFAR-100}} 
& \fontsize{8pt}{1em}\selectfont \bf \textbf{\texttt{TINY-IMAGENET}} \\
\hline
\modelname$_{CE}$ & \quad \textbf{91.99}	& \quad \textbf{69.43}	&  \quad \textbf{48.12}	\\
\modelname$_{NT-Xent}$ & \quad 91.95	& \quad 69.27	&  \quad 47.29	\\
\hline
\end{tabular}
}
\label{tab:acc_compare_loss_variations}
\end{table}

\subsection{Loss function for gradient calculation}\label{subsup:loss_ablation}
\begin{figure}[t!]
    \centering
    \subfigure[\textsc{Cross-Entropy}]{
    \resizebox{.47\textwidth}{!}{
    \includegraphics[width=\textwidth, height=.7\textwidth]{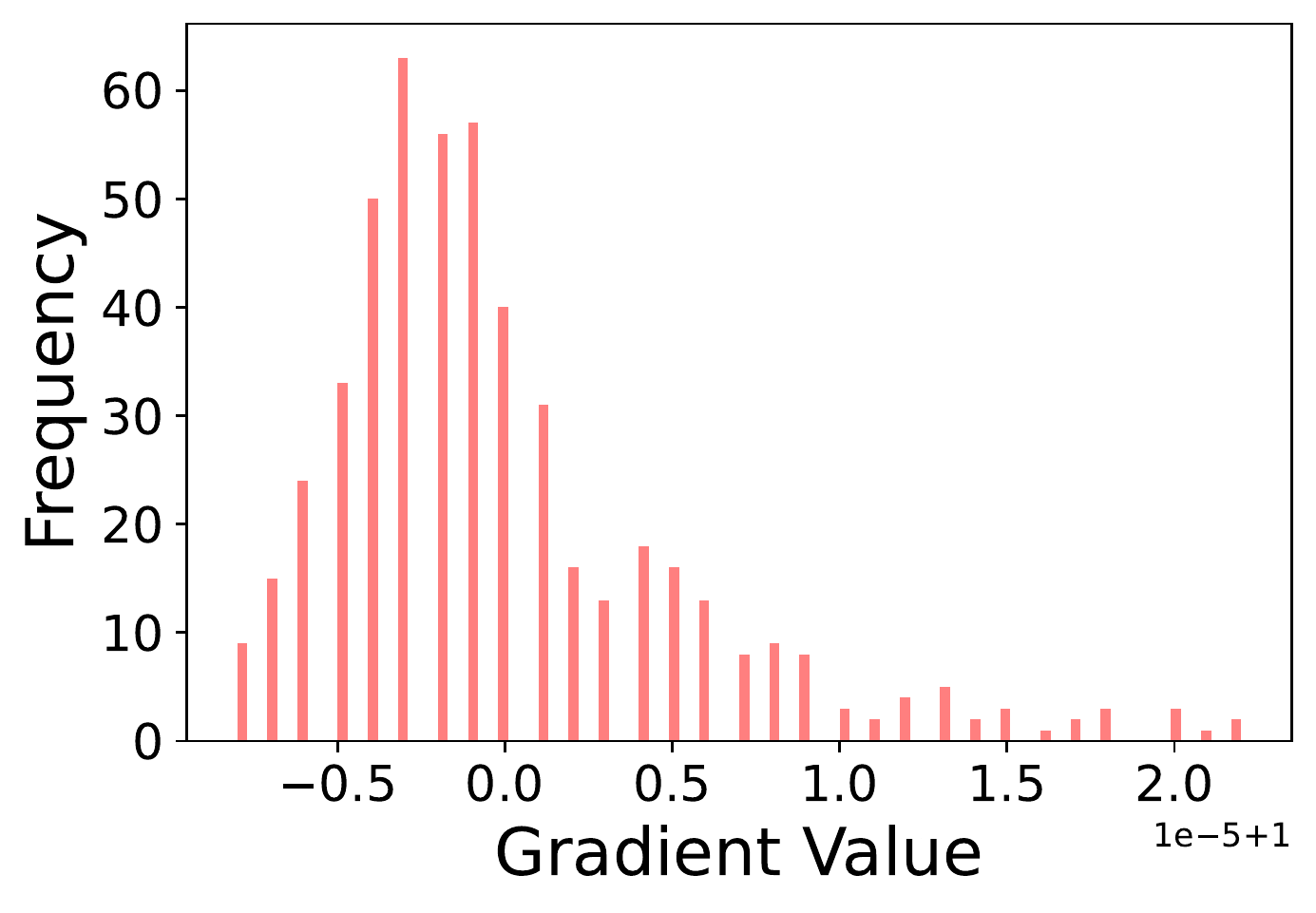}
    } }
    \subfigure[\textsc{NT-Xent}]{
    \resizebox{.47\textwidth}{!}{
    \includegraphics[width=\textwidth, height=.7\textwidth]{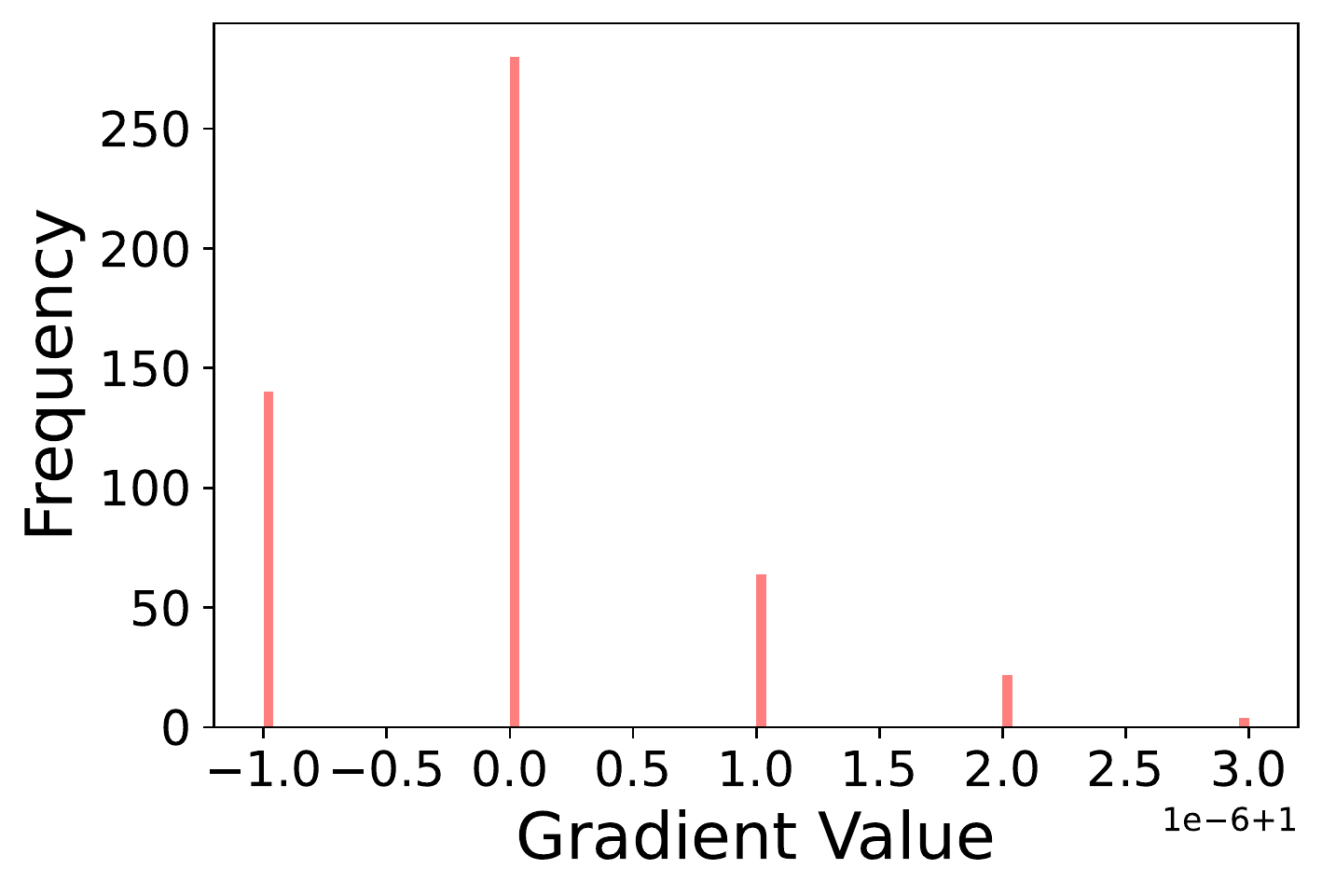}
    } }
    \vspace{-0.1in}
    \caption{Distribution of gradients of negative examples using (a) Cross-Entropy Loss and (b) NT-Xent Loss.}
    \label{fig:loss_comparison}
\end{figure}

We experiment with two types of loss functions for calculating gradients w.r.t. each negative example in Eq. (\ref{eq:gradient}): 1) \textsc{Cross-Entropy} (CE) loss and 2) \textsc{NT-Xent}, \ie the Normalized Temperature-scaled Cross-Entropy loss in Eq. (\ref{eq:nce}). Table~\ref{tab:acc_compare_loss_variations} presents top-$1\%$  linear evaluation accuracy for \modelname trained with both variants and Figure~\ref{fig:loss_comparison} presents the distribution of gradient values for negative examples. We notice that \textsc{CE} gradients exhibit more variance than \textsc{NT-Xent} gradients (Figure~\ref{fig:loss_comparison}), and that the gradient component based on \textsc{CE} results in higher performance than the \textsc{NT-Xent} variation (Table~\ref{tab:acc_compare_loss_variations}). This might allude to the usefulness of pseudo-labeling strategies for tasks with limited labels~\cite{lee2013pseudo,zhai2019s4l}.

\section{Evaluation on \textsc{Imagenet-1k}}
We additionally evaluate \modelname on \textsc{Imagenet-1k}~\cite{deng2009imagenet} using {MoCo-V2}~\cite{chen2020mocov2} as backbone. For this experiment, we make use of the recently released FFCV data loader~\cite{leclerc2022ffcv}. We directly use the MoCo-V2 official implementation. Due to resource constraints, and only for this experiment, we run all compared models for fewer epochs. Table~\ref{tab:imagenet_acc} presents the top-$1\%$ and top-$5\%$ accuracy after training with contrastive loss for 50 epochs, and then fine-tuning for 70 epochs, with \modelname outperforming MoCo-V2. 
\begin{table}[h!]
\newcommand*{\fct}[1]{\multicolumn{1}{>{\columncolor{white}\hspace*{-\tabcolsep}}c}{#1}}
\centering
\caption{Comparison of performance on \textsc{Imagenet-1k}}
\resizebox{0.8\textwidth}{!}{
\begin{tabular}{ll|ll}
\hline
& & \multicolumn{2}{c}{\fontsize{8pt}{1em}\selectfont \texttt{\textbf{Accuracy}}} \\
\fontsize{8pt}{1em}\selectfont \bf Method 
& \quad \fontsize{8pt}{1em}\selectfont \bf \texttt{\textbf{Backbone}} 
& \fontsize{8pt}{1em}\selectfont \bf $~~~~$\texttt{\textbf{Top-1\%}} 
& \fontsize{8pt}{1em}\selectfont \bf $~~~~$\texttt{\textbf{Top-5\%}} \\
\hline
HCL~\citep{robinson2020hard} &  \quad SimCLR	& \quad 38.28 &	\quad 61.85 \\

MoCo-V2~\citep{chen2020mocov2} &  \quad MoCo	&  \quad 38.99 &  \quad	62.92 \\
\hline
\rowcolor{mygray} \cellcolor{white} \bf \modelname	& \cellcolor{white} \quad MoCo	& \quad \textbf{39.71}\color{darkergreen}$^{\bf\uparrow0.72}$ & \quad \textbf{63.63}\color{darkergreen}$^{\bf\uparrow0.71}$ \\
\hline
\end{tabular}
}
\label{tab:imagenet_acc}
\end{table}

\section{Limitations}\label{sec:limitation}
Even though gradient-based uncertainty reduces the importance of most similar negative examples for which the model has low uncertainty, it fails to reduce the importance of most similar negative examples with high uncertainty (closer to the decision boundary). However, qualitative analysis shows that the occurrence of such a situation is so infrequent that it barely affects model training.
\section{Qualitative Evaluation}
\begin{figure}[t!]
    \centering
    \subfigure[]{\resizebox{.48\linewidth}{!}{\includegraphics[width=.7\linewidth]{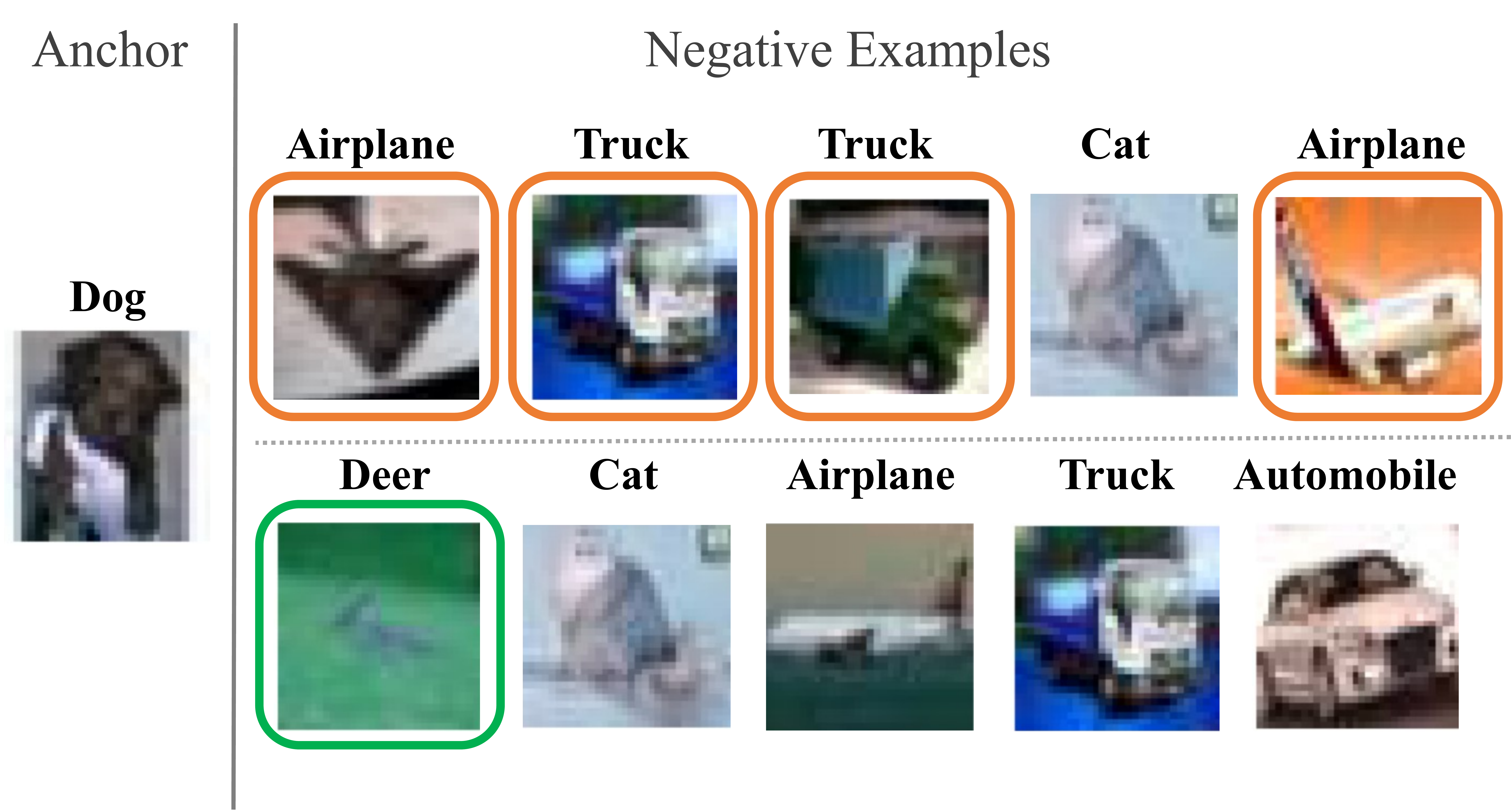}} }
    \rulesep
    \subfigure[]{\resizebox{.48\linewidth}{!}{\includegraphics[width=.7\linewidth]{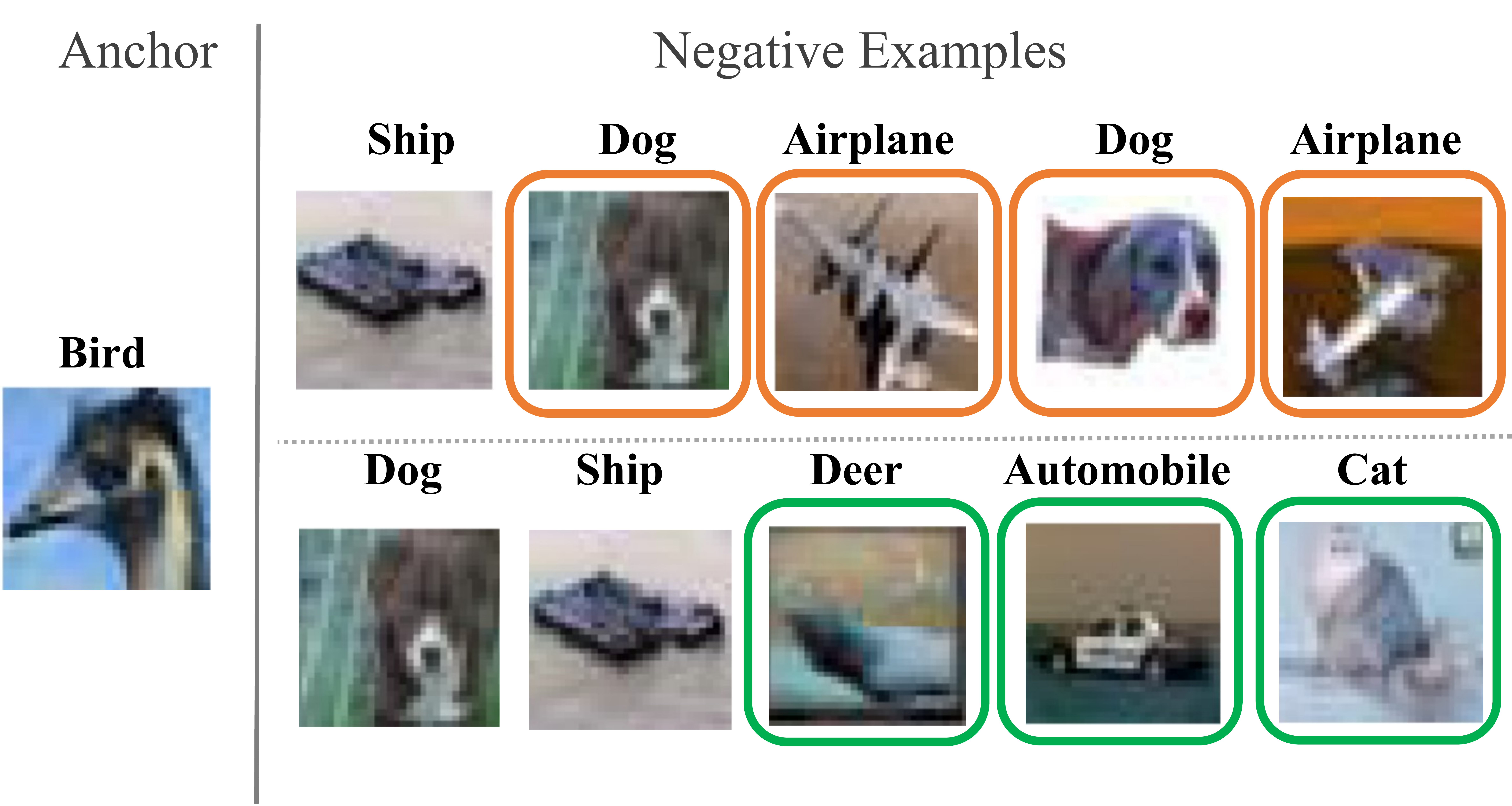}} }
    \subfigure[]{\resizebox{.48\linewidth}{!}{\includegraphics[width=.7\linewidth]{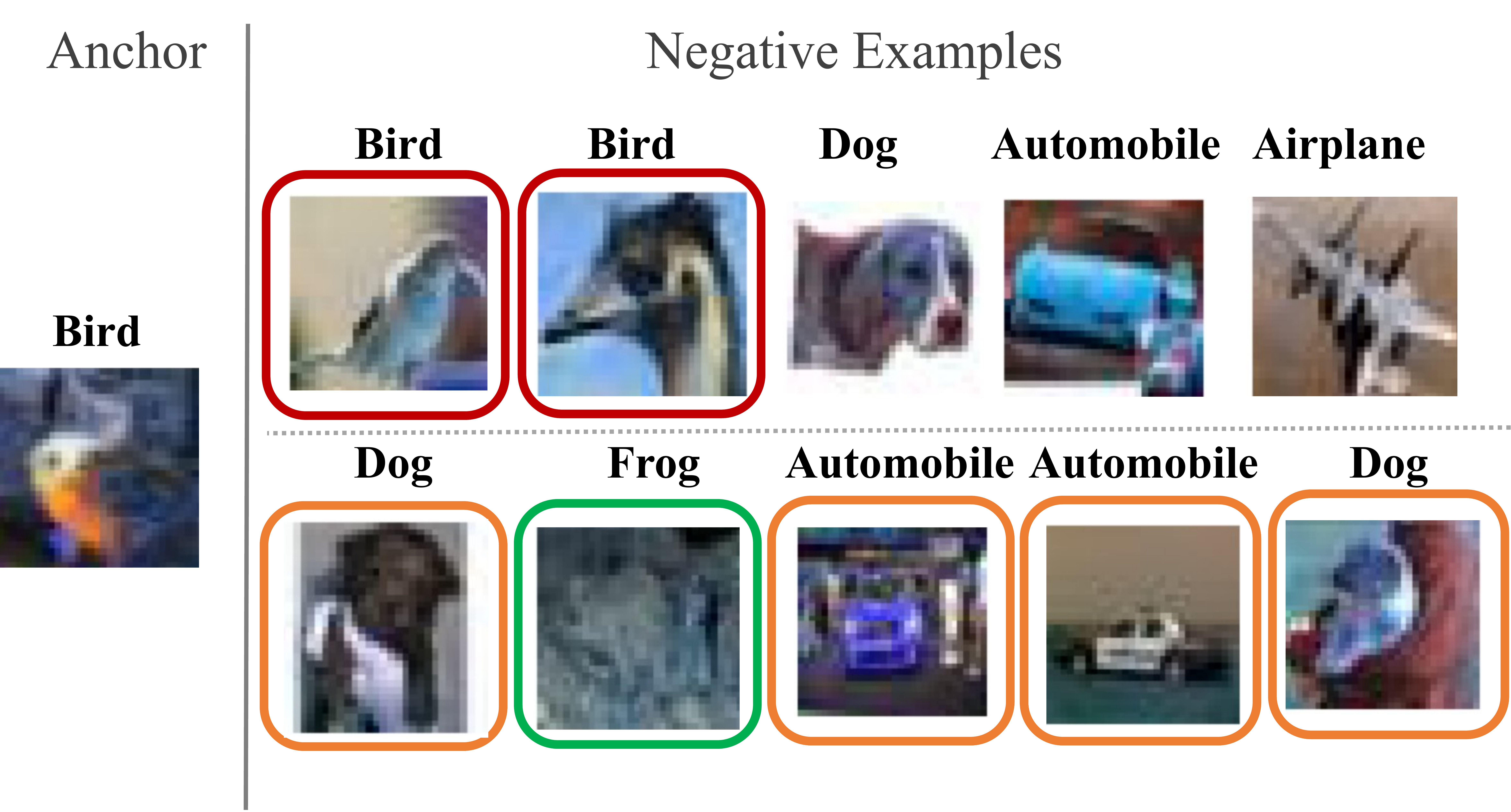}} }
    \rulesep
    \subfigure[]{\resizebox{.48\linewidth}{!}{\includegraphics[width=.7\linewidth]{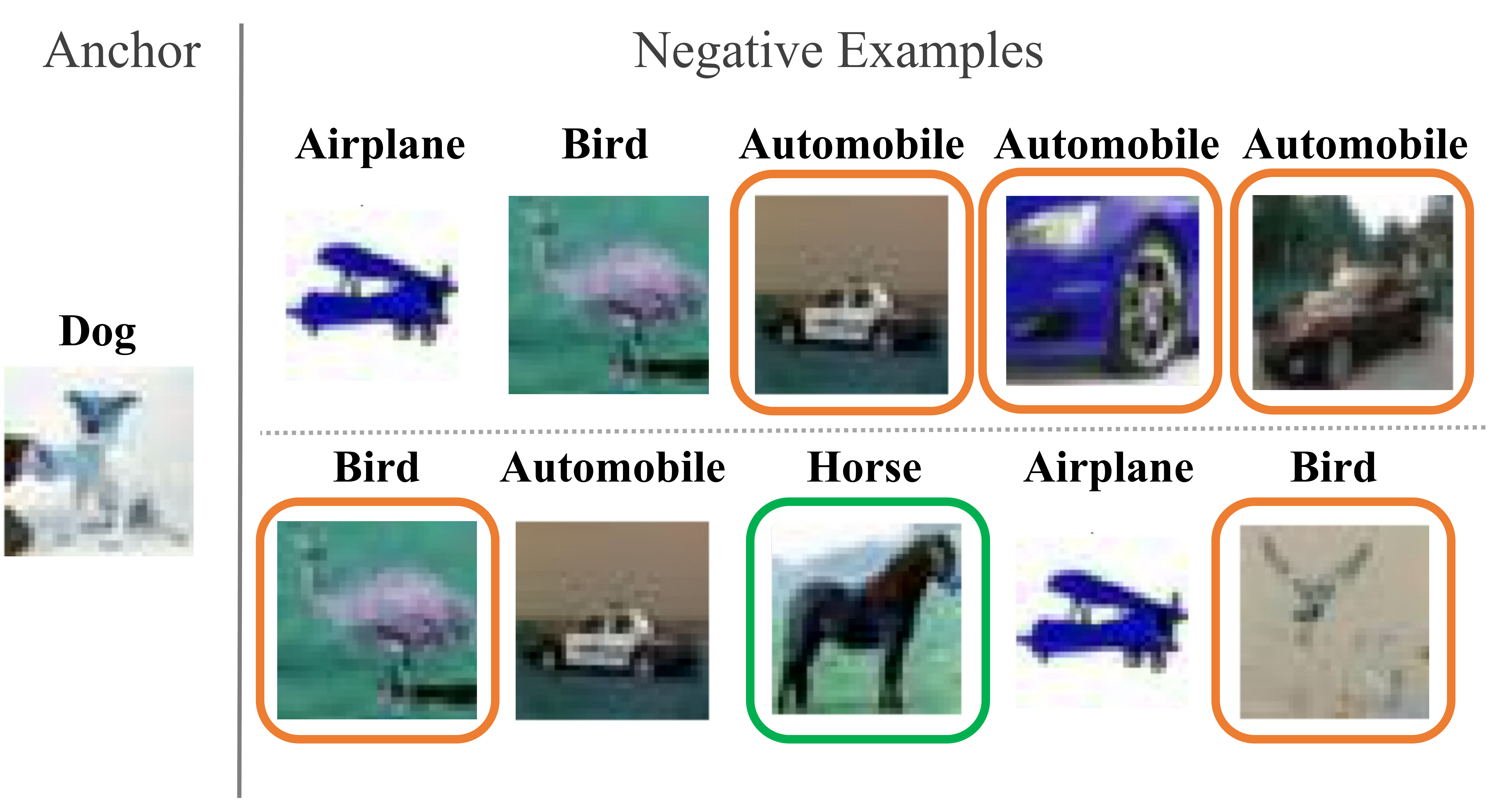}} }
    \subfigure[]{\resizebox{.48\linewidth}{!}{\includegraphics[width=.7\linewidth]{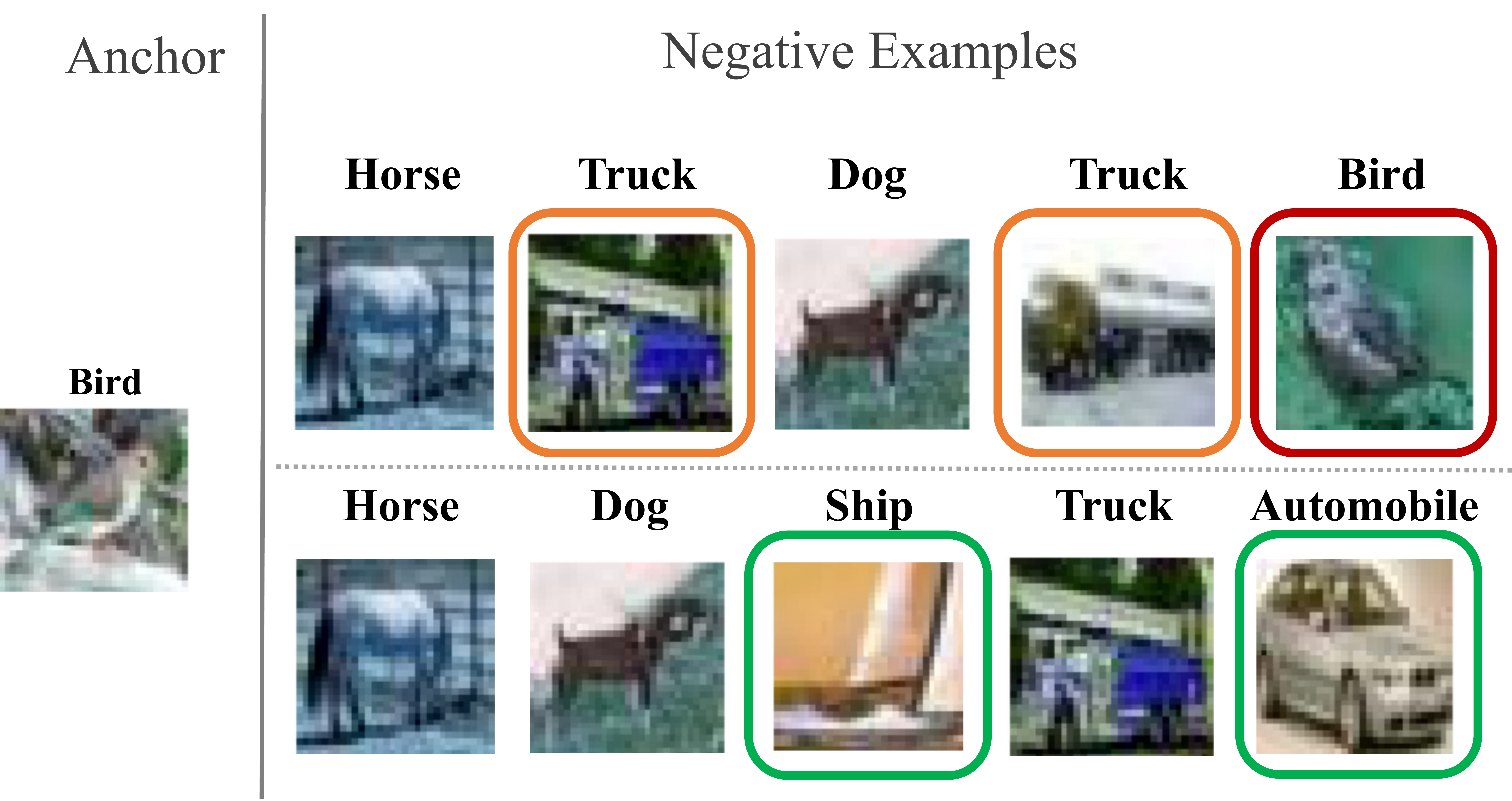}} }
    \rulesep
    \subfigure[]{\resizebox{.48\linewidth}{!}{\includegraphics[width=.7\linewidth]{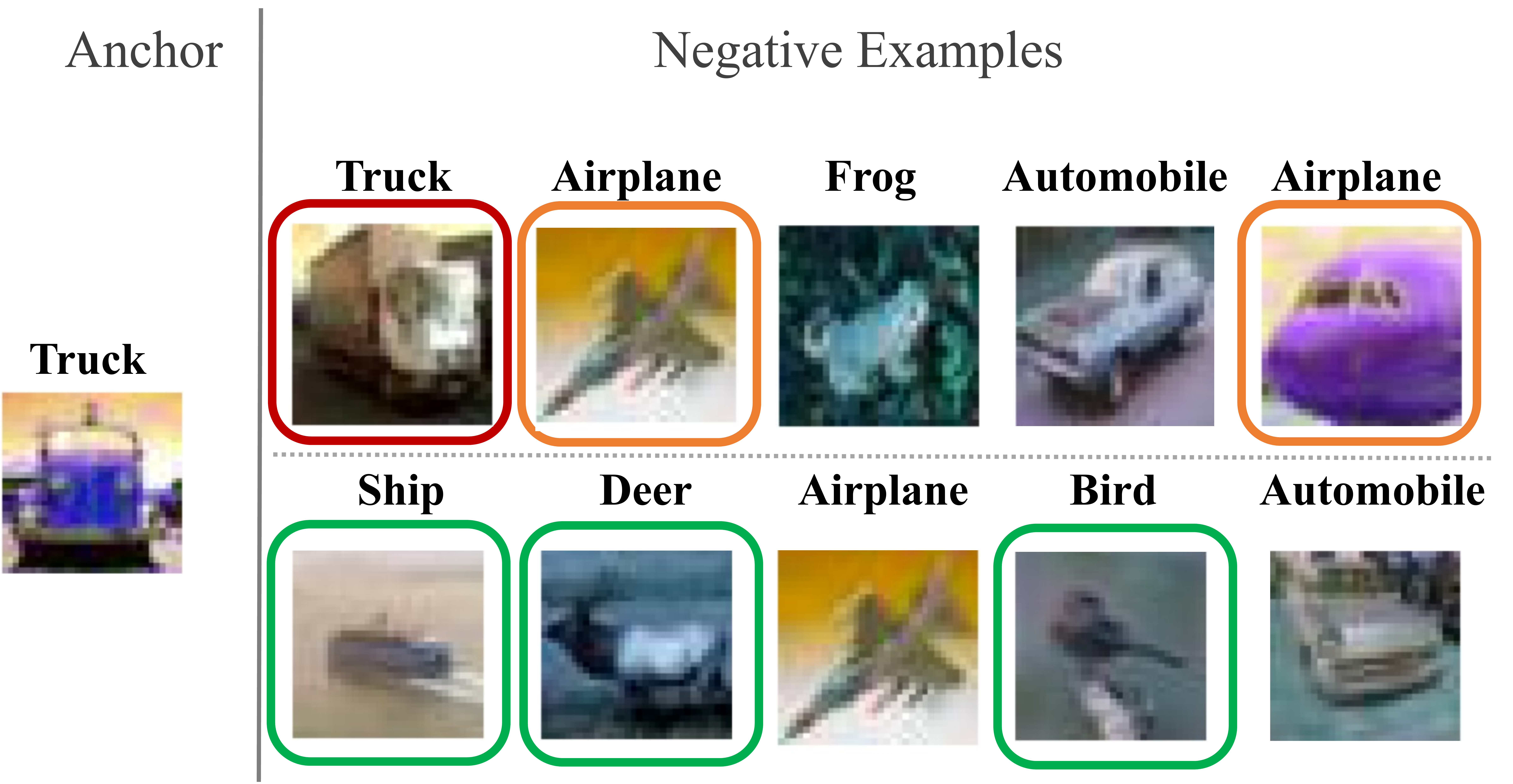}} }
    \subfigure[]{\resizebox{.48\linewidth}{!}{\includegraphics[width=.7\linewidth]{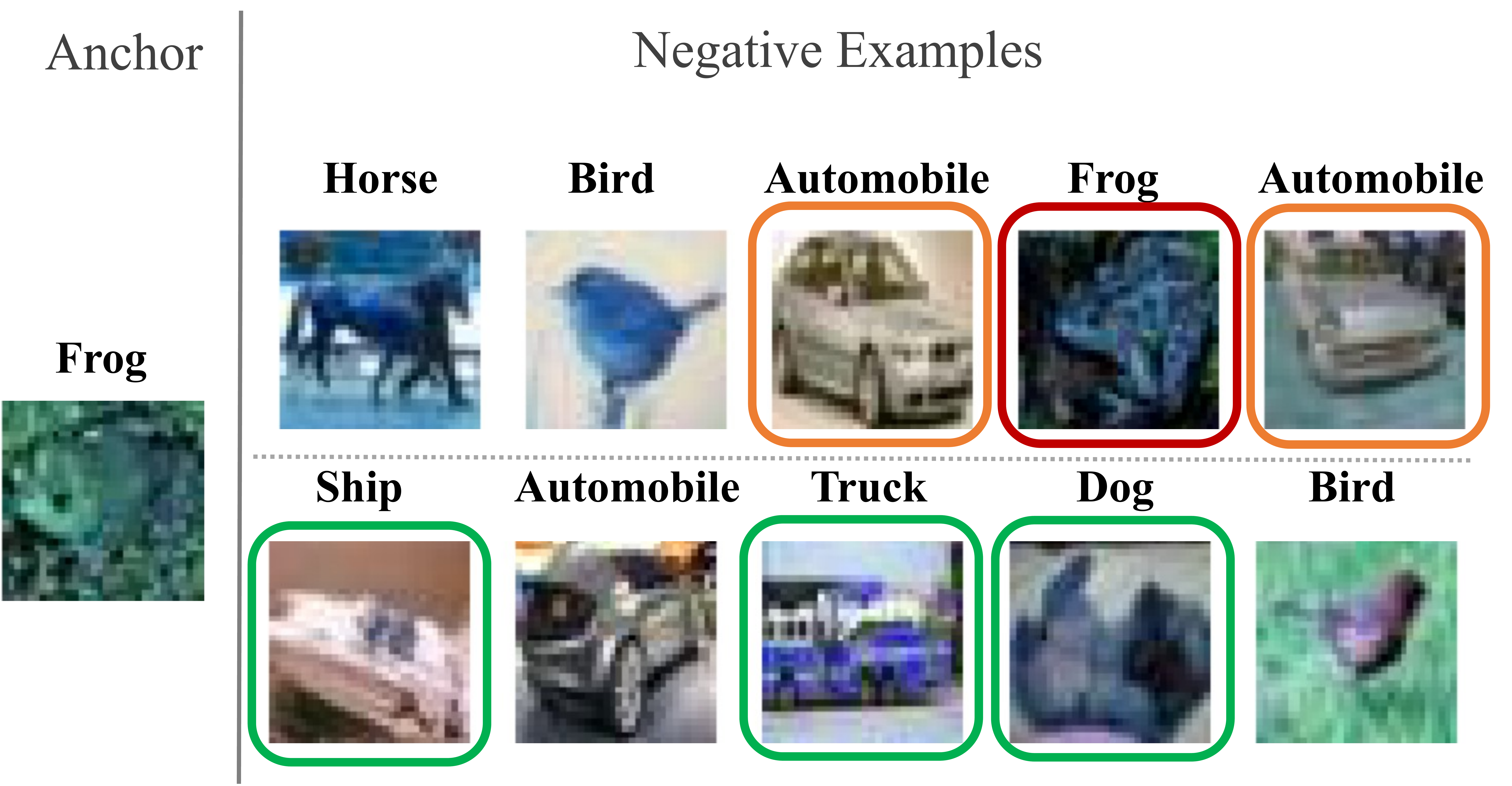}} }
    \rulesep
    \subfigure[]{\resizebox{.48\linewidth}{!}{\includegraphics[width=.7\linewidth]{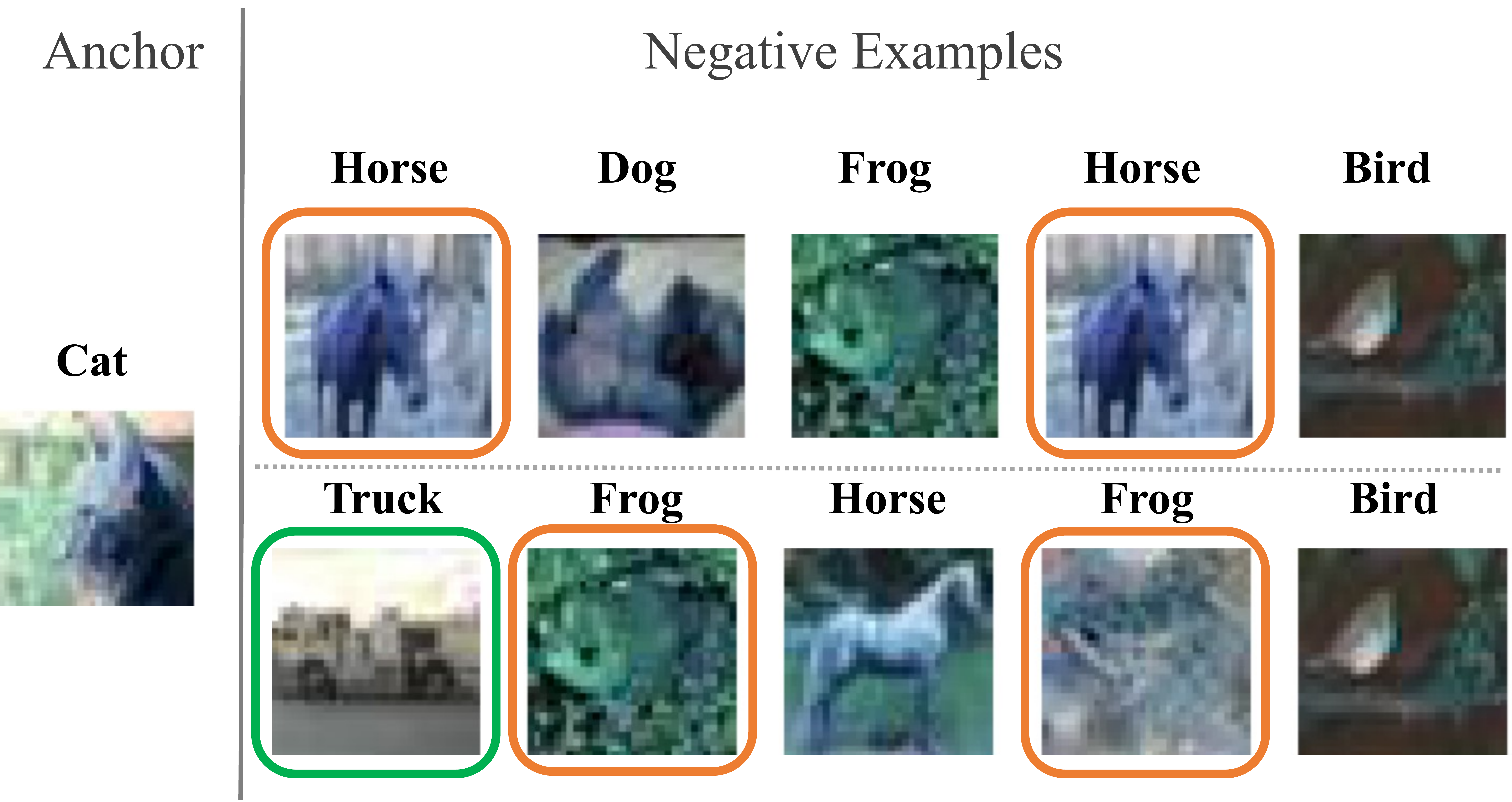}} }
    \caption{Qualitative evaluations on \texttt{CIFAR-10}, best viewed in color. Top Row: HCL~\citep{robinson2020hard} has sampled negative examples of the same class as the anchor as a negative example (\textcolor{red}{\textbf{red}} bbox) or selected examples of the same class multiple times (\textcolor{orange}{\textbf{orange}} bboxes). Bottom Row: \modelname samples more diverse true negative examples and avoids false negative examples (\textcolor{green}{\textbf{green}} bboxes). }
    \label{fig:cifar10_qualitative_examples}
\end{figure}

\begin{figure}[t!]
    \centering
    \subfigure[]{\resizebox{.48\linewidth}{!}{\includegraphics[width=.7\linewidth]{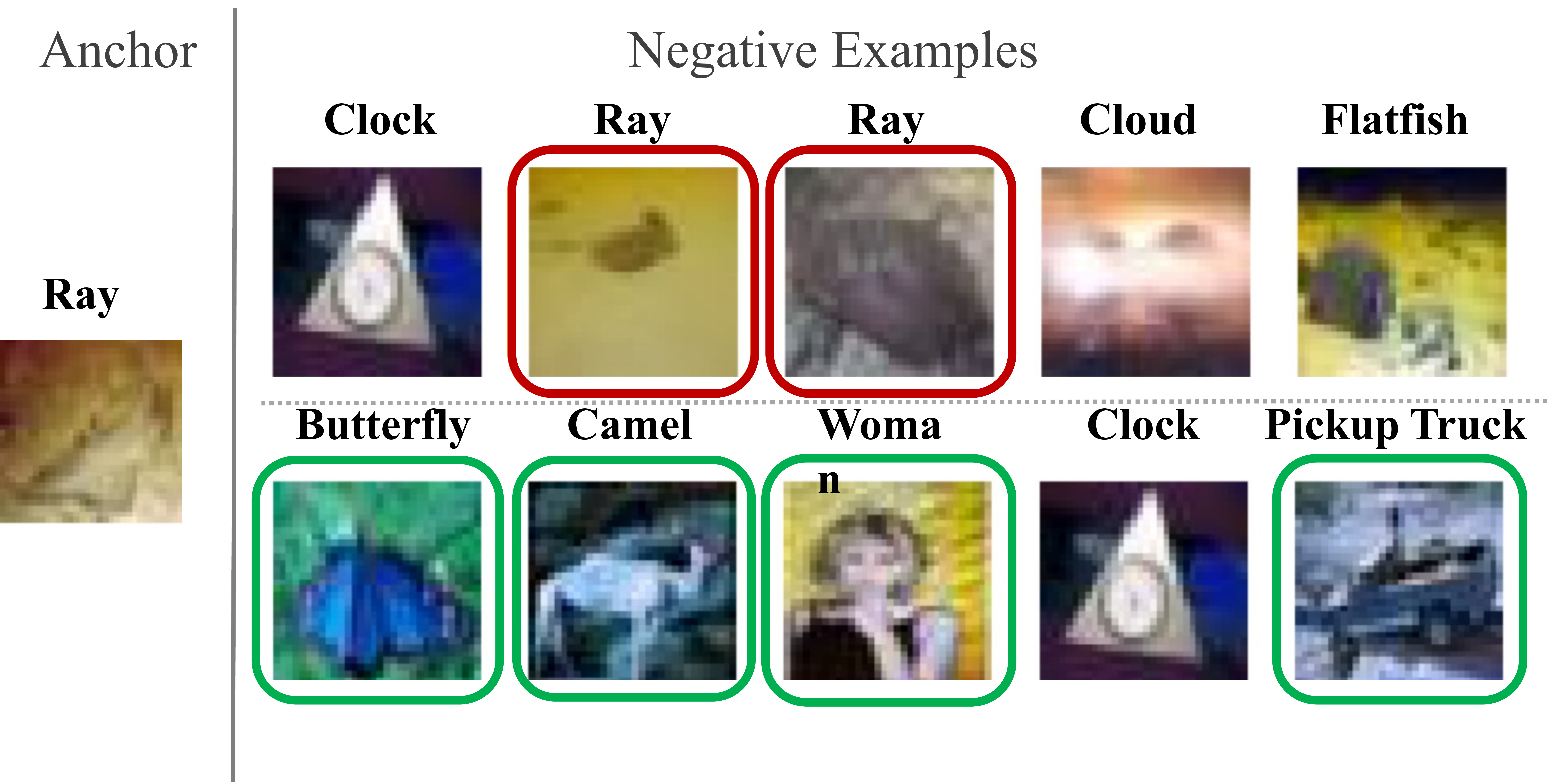}} }
    \rulesep
    \subfigure[]{\resizebox{.48\linewidth}{!}{\includegraphics[width=.7\linewidth]{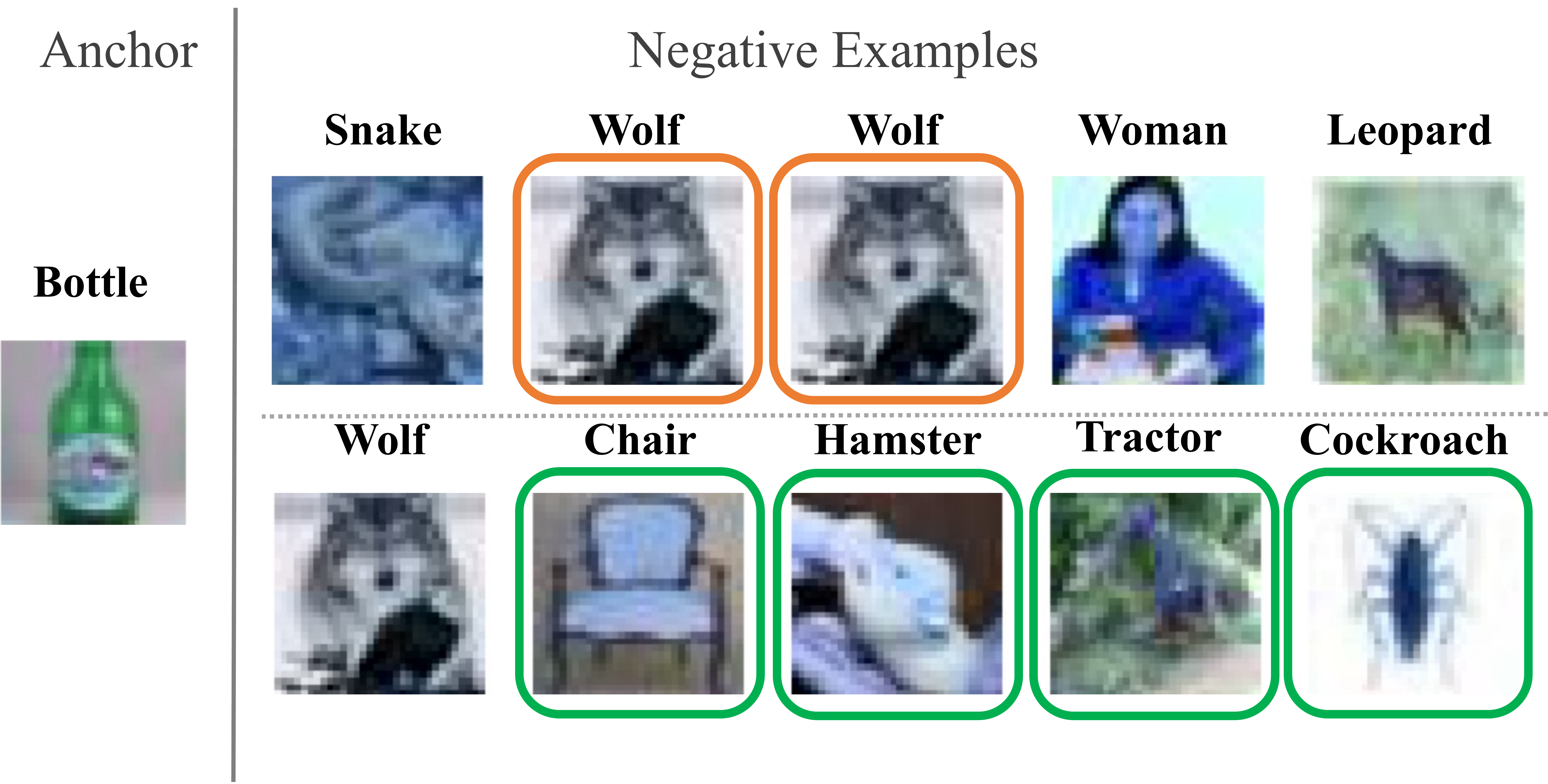}} }
    \subfigure[]{\resizebox{.48\linewidth}{!}{\includegraphics[width=.7\linewidth]{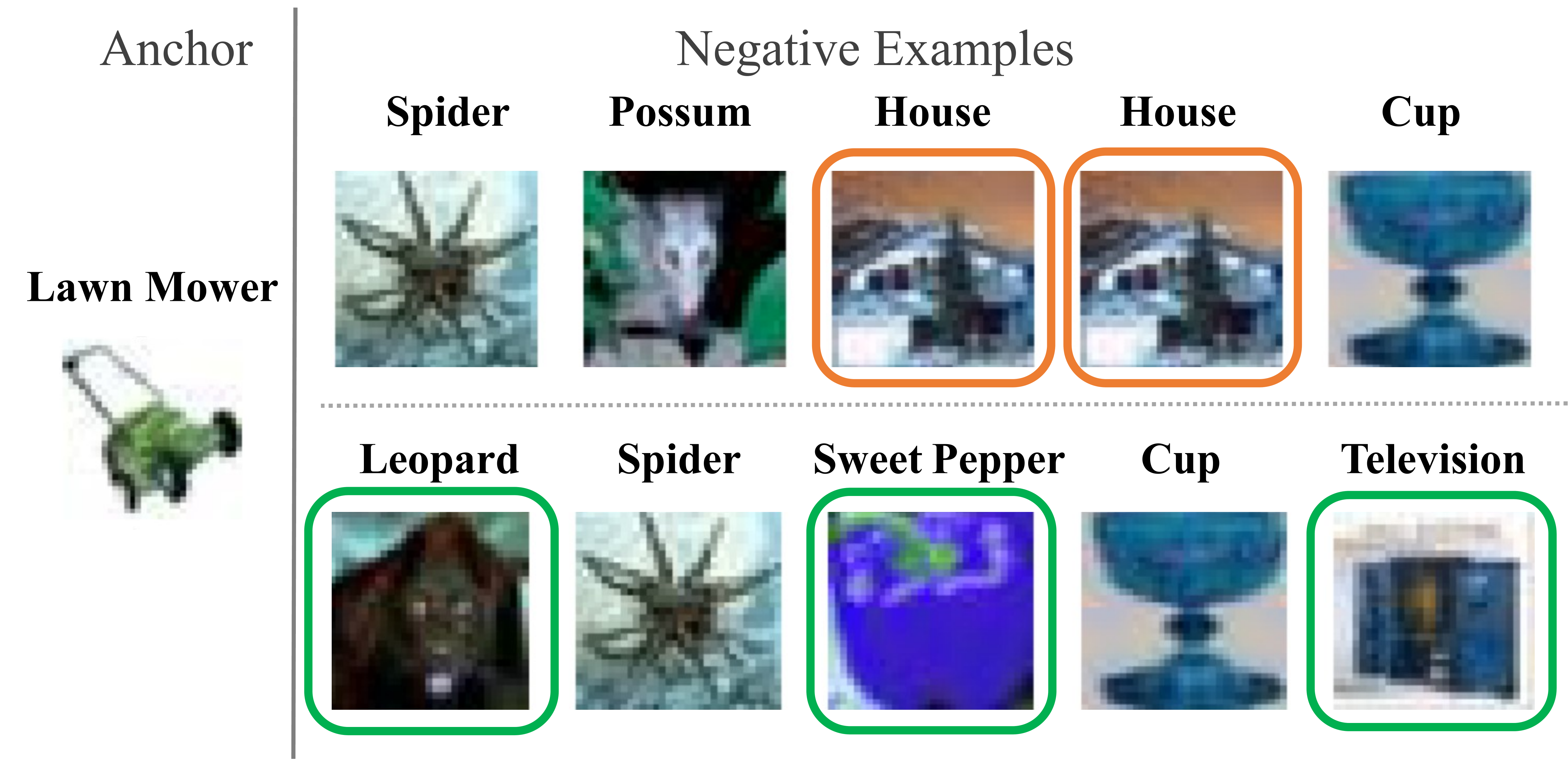}} }
    \rulesep
    \subfigure[]{\resizebox{.48\linewidth}{!}{\includegraphics[width=.7\linewidth]{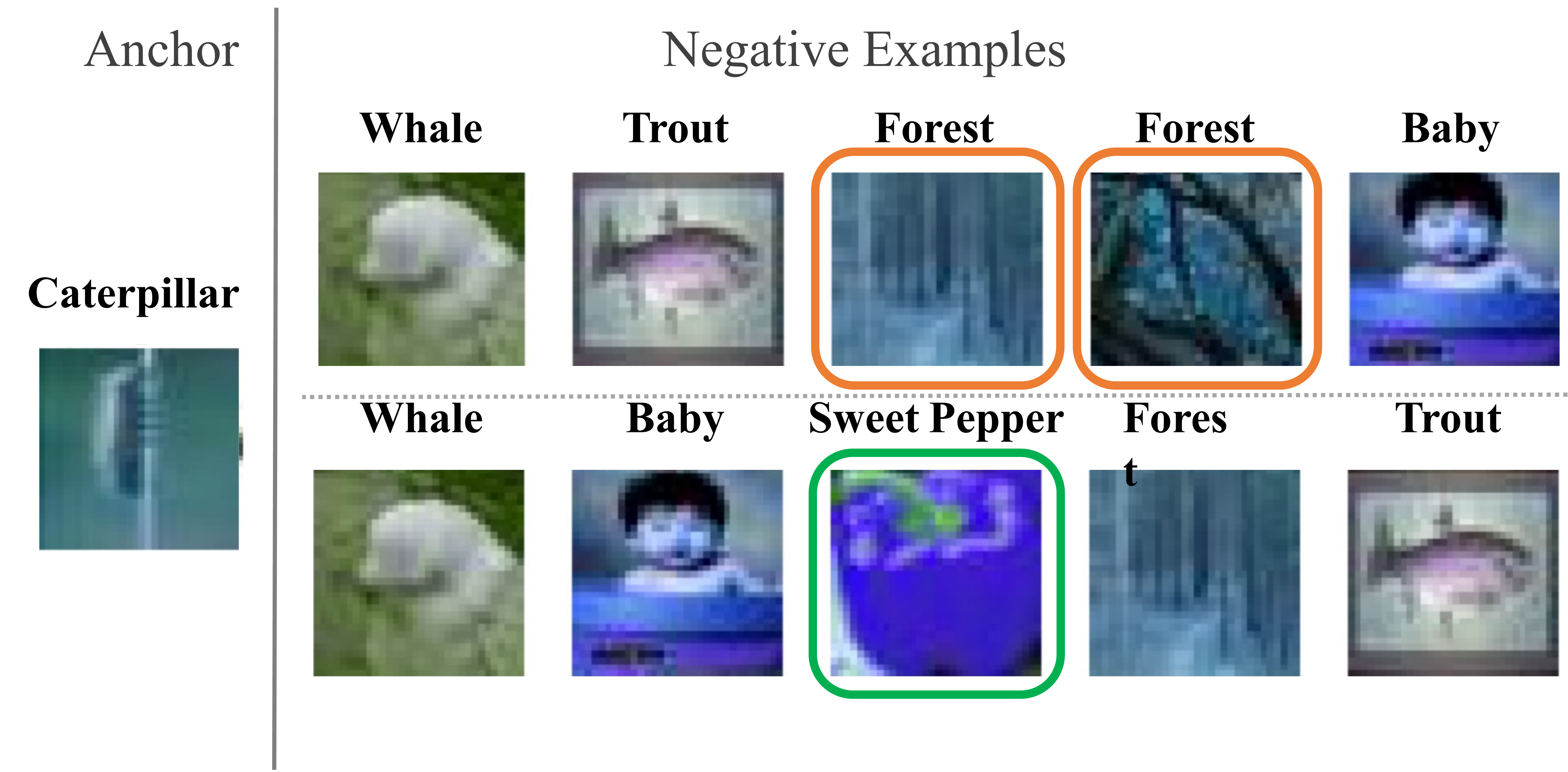}} }
    \subfigure[]{\resizebox{.48\linewidth}{!}{\includegraphics[width=.7\linewidth]{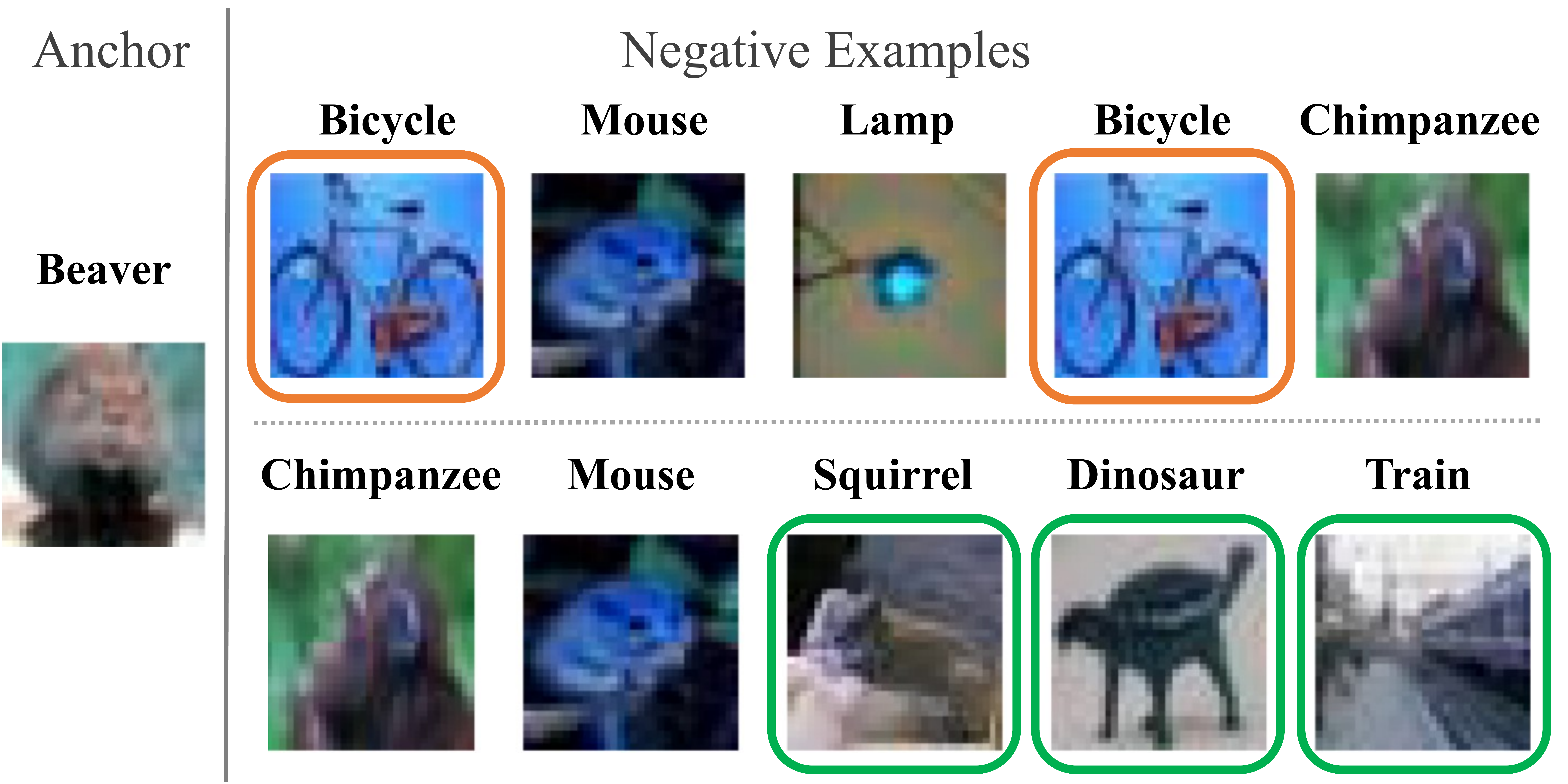}} }
    \rulesep
    \subfigure[]{\resizebox{.48\linewidth}{!}{\includegraphics[width=.7\linewidth]{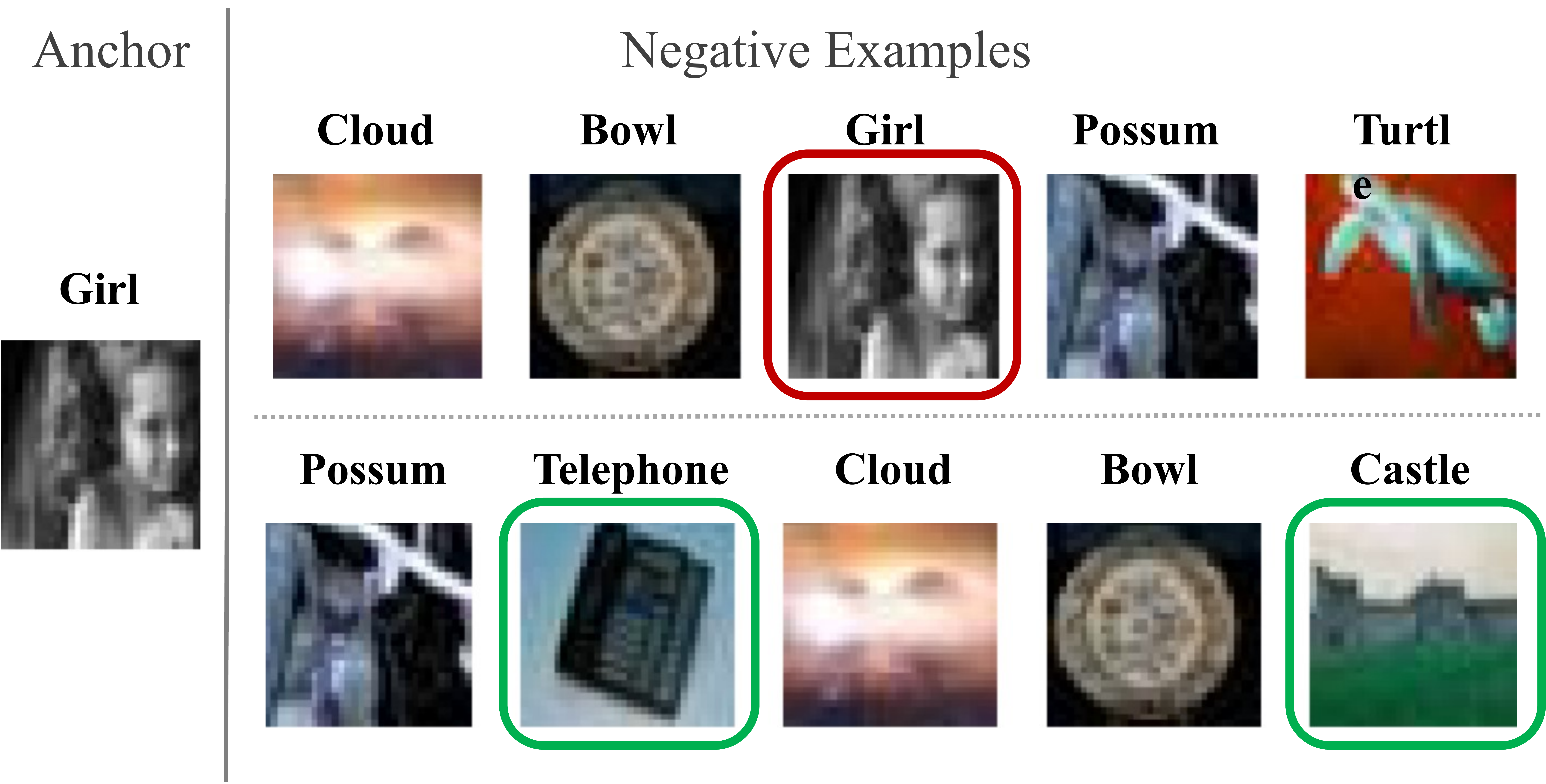}} }
    \subfigure[]{\resizebox{.48\linewidth}{!}{\includegraphics[width=.7\linewidth]{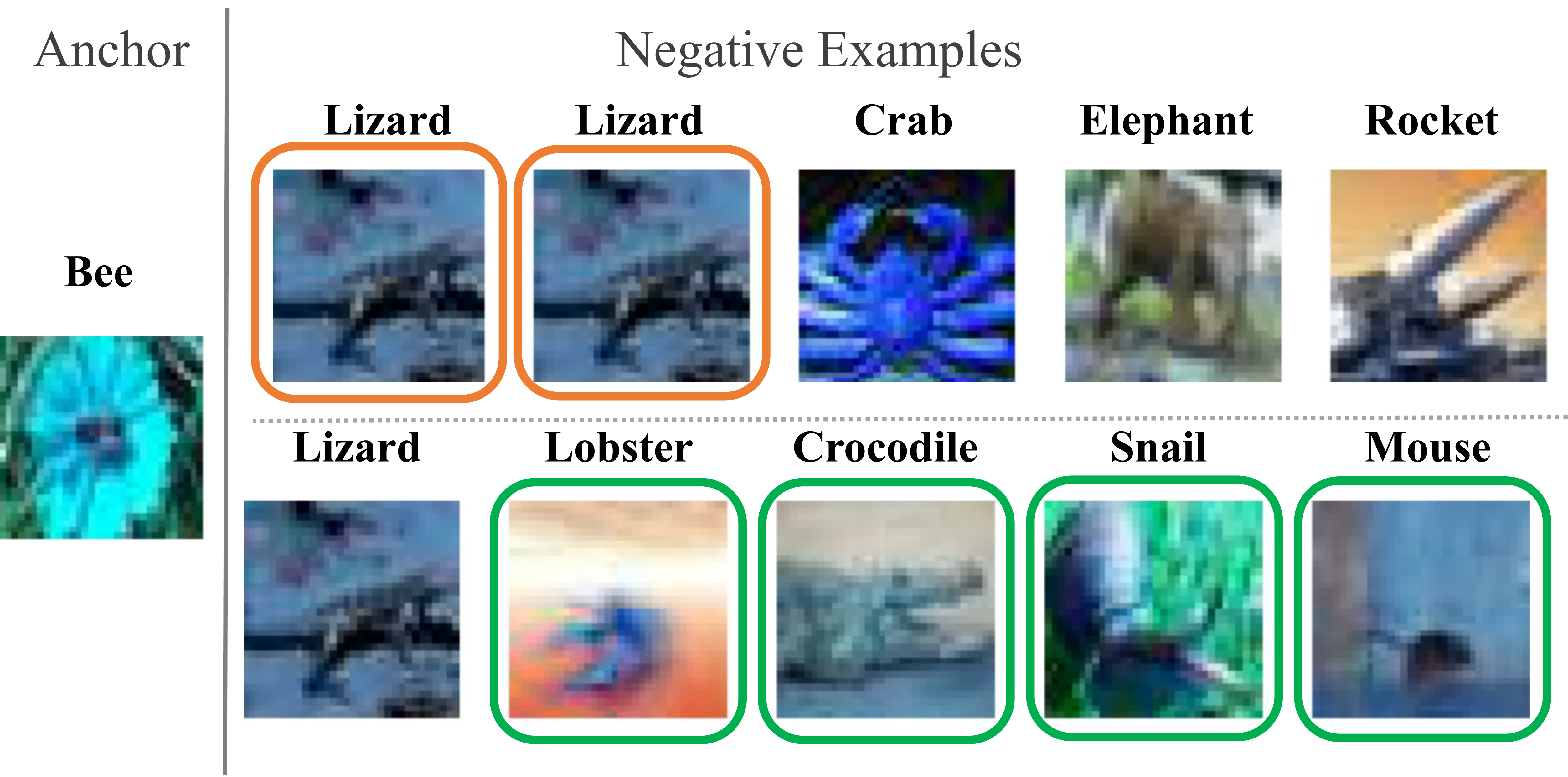}} }
    \rulesep
    \subfigure[]{\resizebox{.48\linewidth}{!}{\includegraphics[width=.7\linewidth]{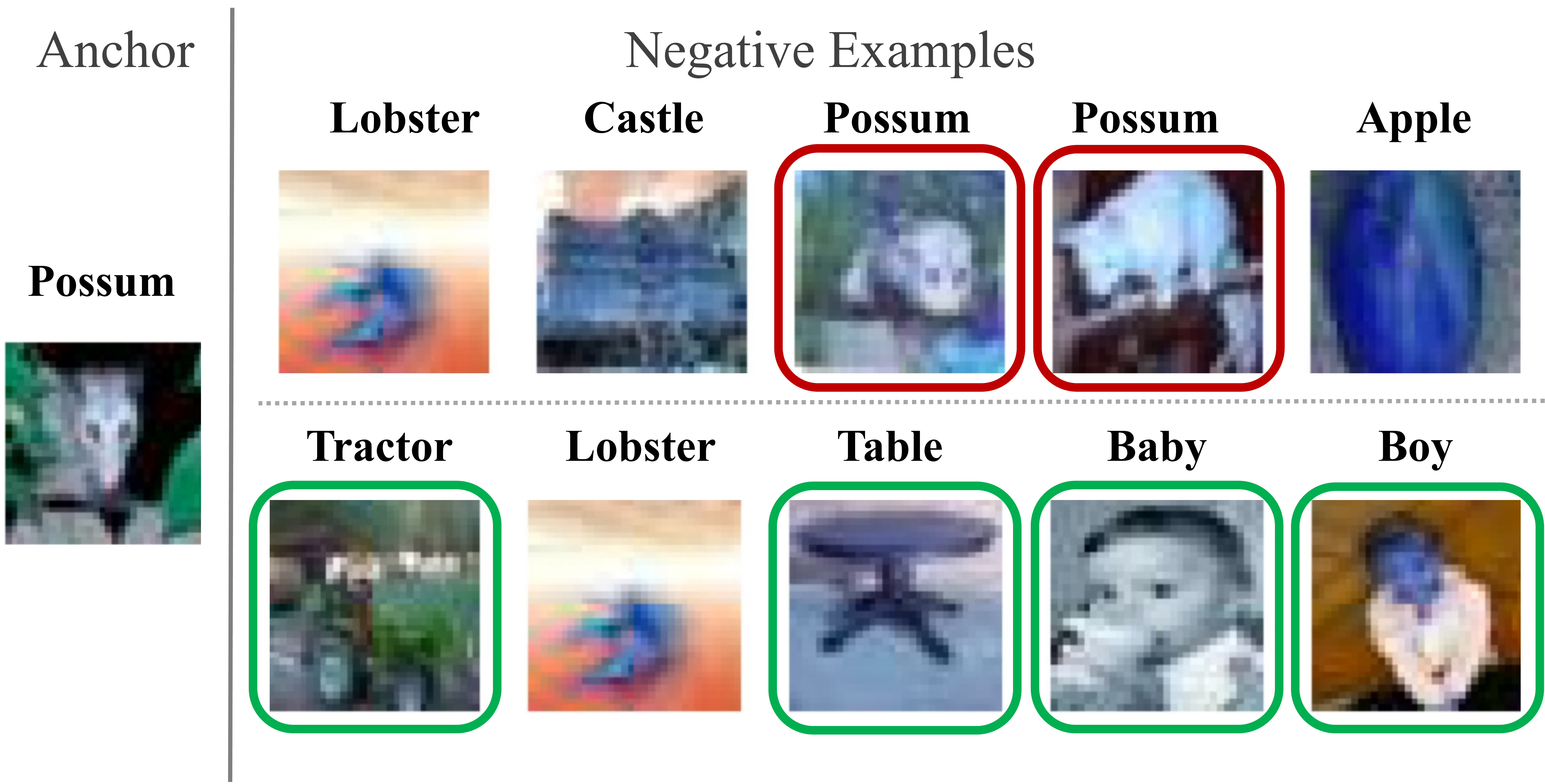}} }
    \caption{Qualitative evaluations on \texttt{CIFAR-100}, best viewed in color. Top Row: HCL~\citep{robinson2020hard} has sampled negative examples of the same class as the anchor as a negative example (\textcolor{red}{\textbf{red}} bbox) or selected examples of the same class multiple times (\textcolor{orange}{\textbf{orange}} bboxes). Bottom Row: \modelname samples more diverse true negative examples and avoids false negative examples (\textcolor{green}{\textbf{green}} bboxes). }
    \label{fig:cifar100_qualitative_examples}
\end{figure}

\begin{figure}[t!]
    \centering
    \subfigure[]{\resizebox{.9\linewidth}{!}{\includegraphics[width=.7\linewidth]{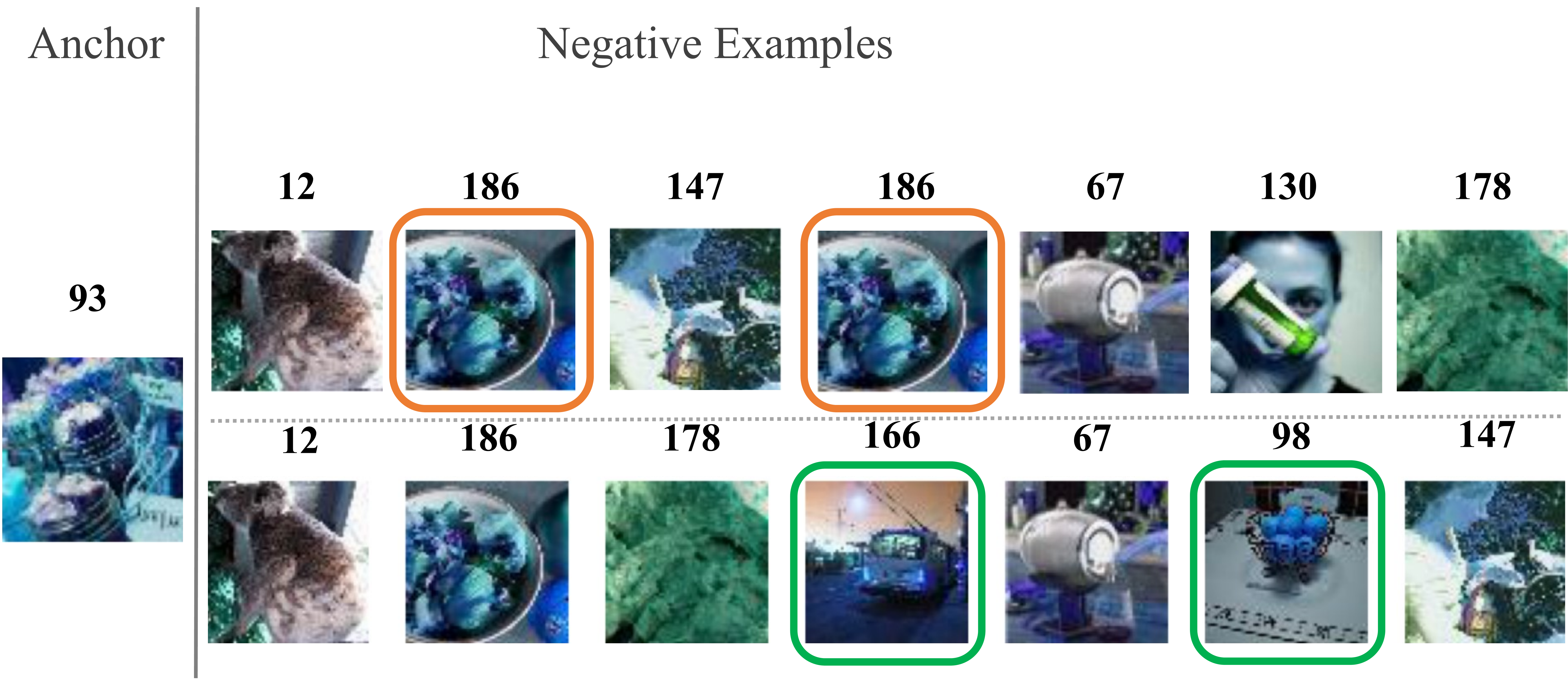}} }
    \subfigure[]{\resizebox{.9\linewidth}{!}{\includegraphics[width=.7\linewidth]{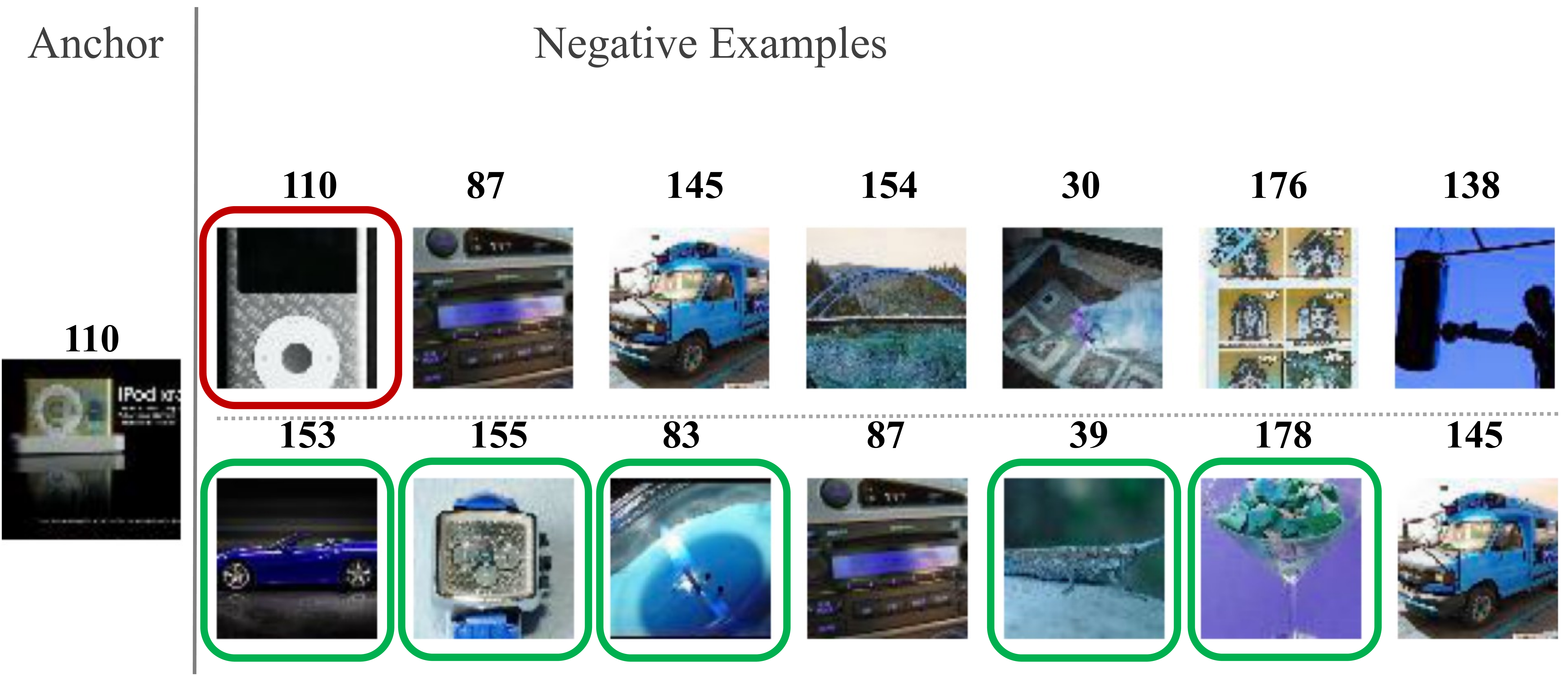}} }
    \subfigure[]{\resizebox{.9\linewidth}{!}{\includegraphics[width=.7\linewidth]{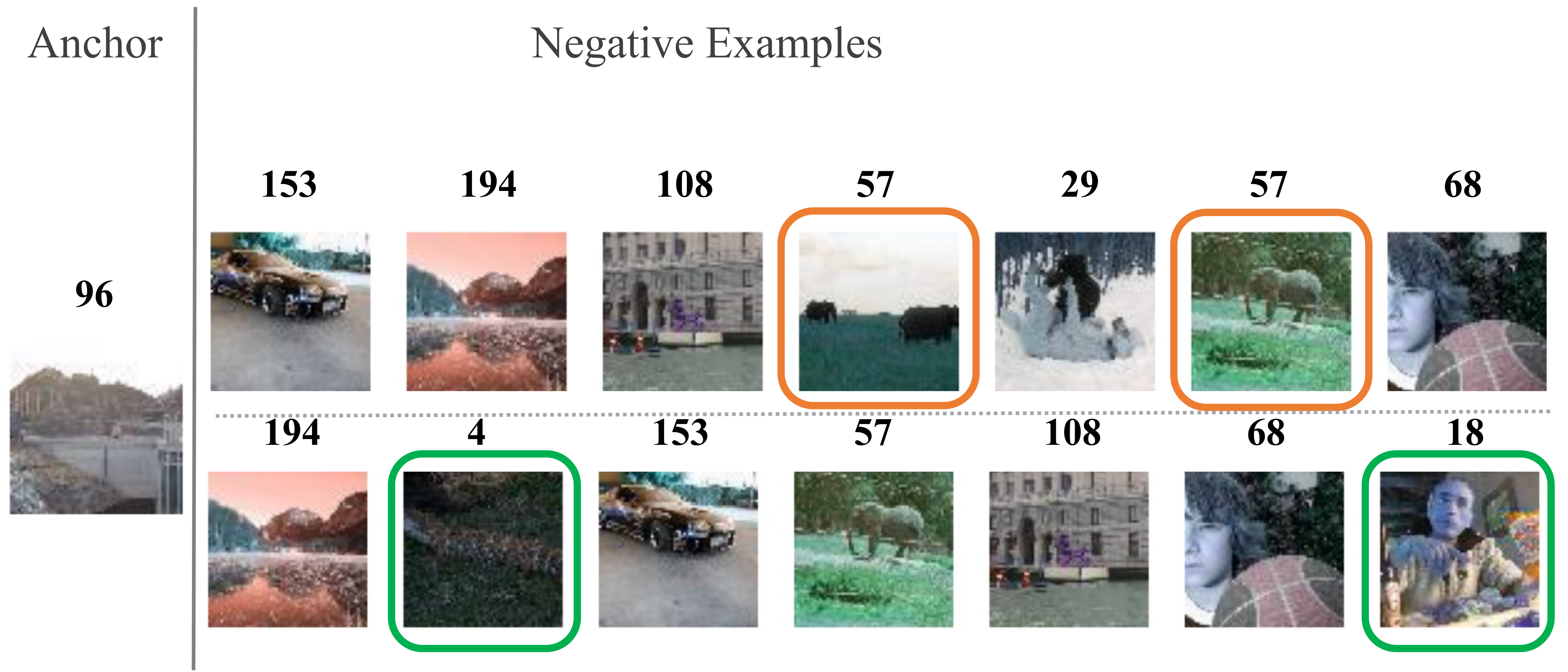}} }
    
\caption{Qualitative evaluations on \texttt{TINY-IMAGENET}, best viewed in color. Top Row: HCL~\citep{robinson2020hard} has sampled negative examples of the same class as the anchor as a negative example (\textcolor{red}{\textbf{red}} bbox) or selected examples of the same class multiple times (\textcolor{orange}{\textbf{orange}} bboxes). Bottom Row: \modelname samples more diverse true negative examples and avoids false negative examples (\textcolor{green}{\textbf{green}} bboxes). }
\label{fig:qualitative_evaluation_tiny_imagenet_1}
\end{figure}

\begin{figure}[t!]
    \centering
    \subfigure[]{\resizebox{.9\linewidth}{!}{\includegraphics[width=.7\linewidth]{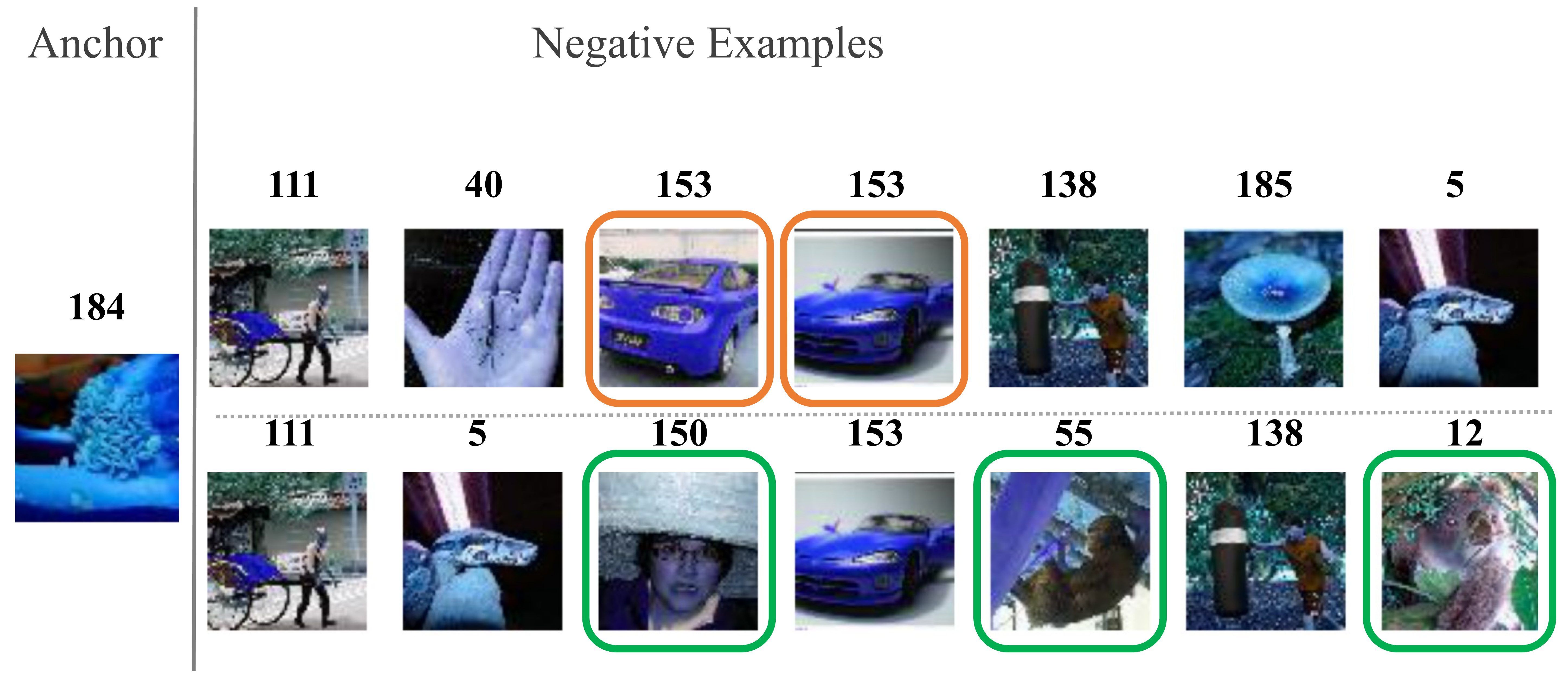}} }
    \subfigure[]{\resizebox{.9\linewidth}{!}{\includegraphics[width=.7\linewidth]{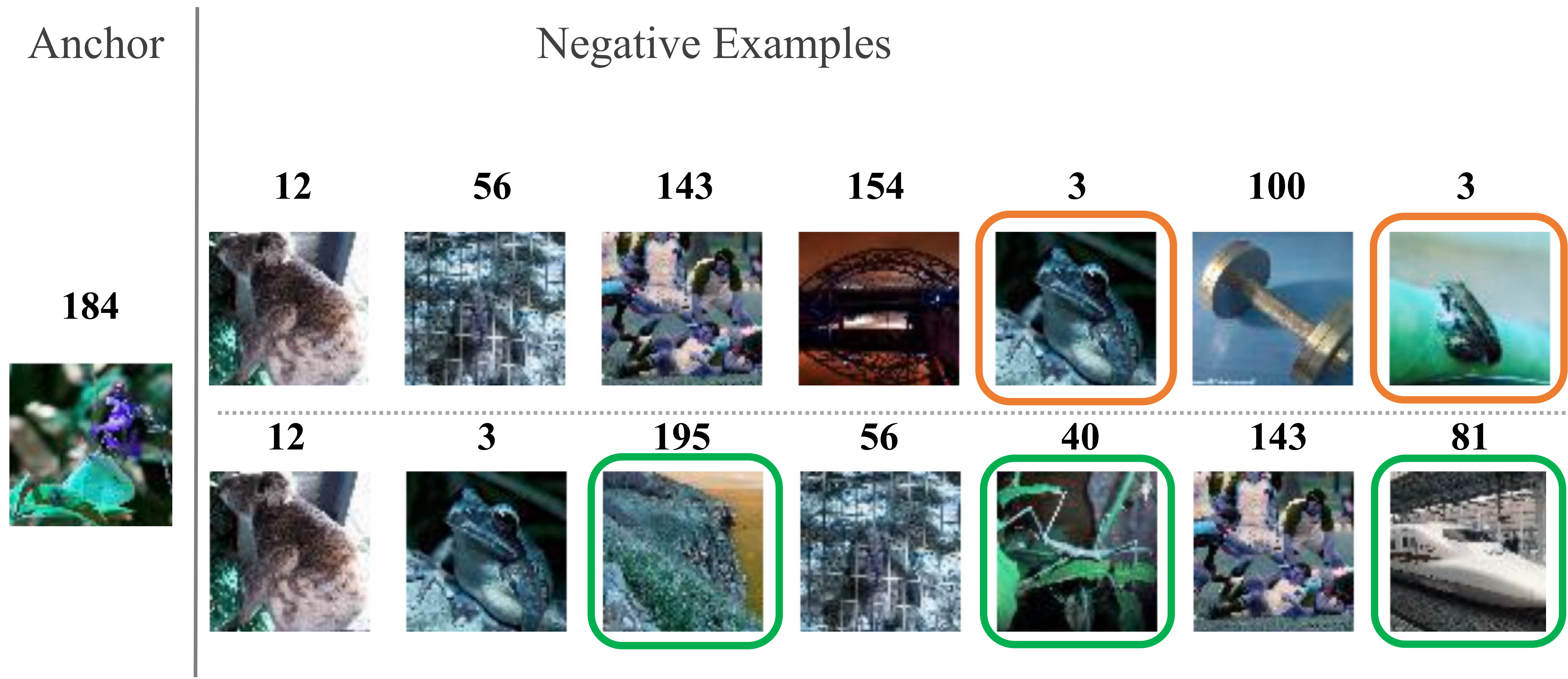}} }
    \subfigure[]{\resizebox{.9\linewidth}{!}{\includegraphics[width=.7\linewidth]{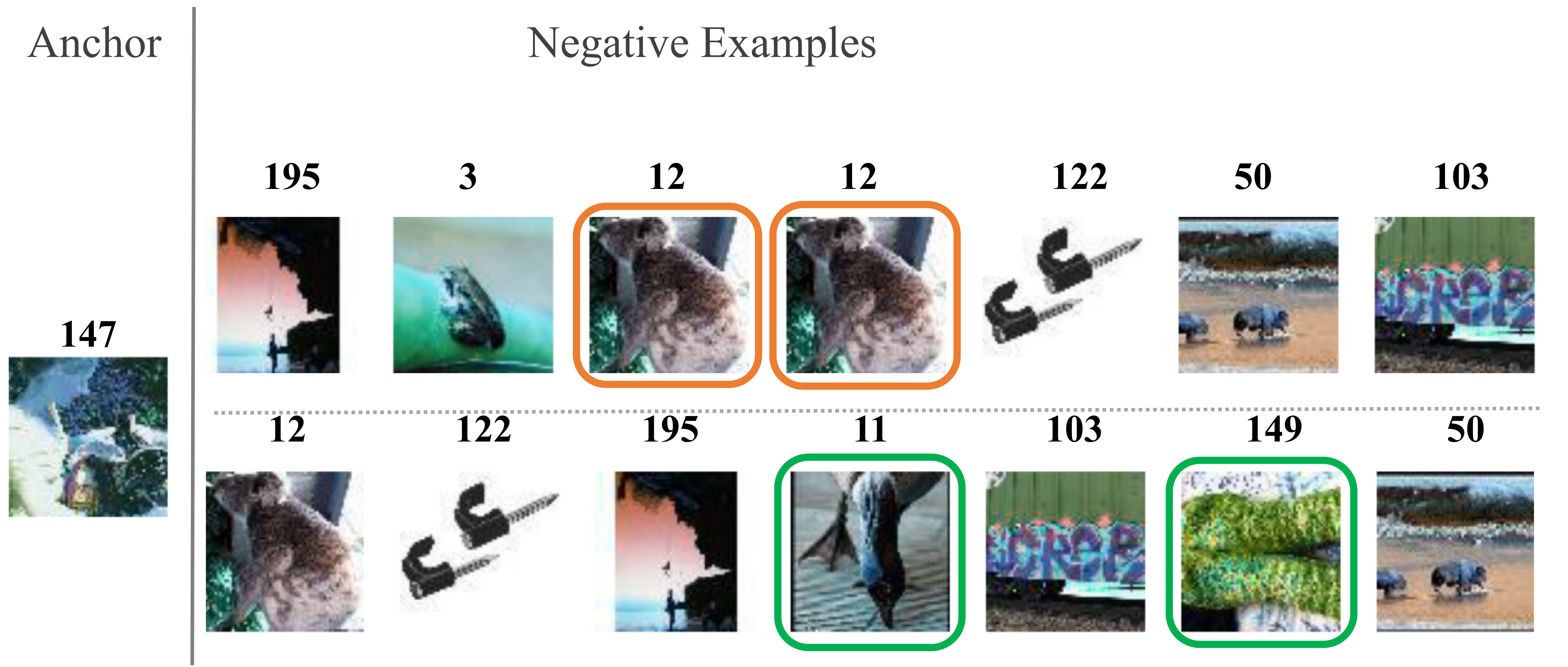}} }
\caption{Qualitative evaluations on \texttt{TINY-IMAGENET}, best viewed in color. Top Row: HCL~\citep{robinson2020hard} has sampled negative examples of the same class as the anchor as a negative example (\textcolor{red}{\textbf{red}} bbox) or selected examples of the same class multiple times (\textcolor{orange}{\textbf{orange}} bboxes). Bottom Row: \modelname samples more diverse true negative examples and avoids false negative examples (\textcolor{green}{\textbf{green}} bboxes). }
\label{fig:qualitative_evaluation_tiny_imagenet_2}
\end{figure}

We present additional qualitative results for \modelname and HCL~\cite{robinson2020hard} for \texttt{CIFAR-10}, \texttt{CIFAR-100} and \texttt{TINY-IMAGENET}.  Figures~\ref{fig:cifar10_qualitative_examples} and~\ref{fig:cifar100_qualitative_examples} depict the five most important negative examples sampled by {HCL} (top row) and \modelname (bottom row), for eight anchor examples from the \texttt{CIFAR-10} and \texttt{CIFAR-100} datasets, respectively. Similarly, Figures~\ref{fig:qualitative_evaluation_tiny_imagenet_1} and~\ref{fig:qualitative_evaluation_tiny_imagenet_2} depict the seven most important negative examples sampled by {HCL} (top row) and \modelname (bottom row), for each of the six anchor examples from the \texttt{TINY-IMAGENET} dataset, observing in total consistent qualitative improvements for more than 22 anchor examples across all datasets. 

\textbf{\texttt{CIFAR-10}:} In Figure~\ref{fig:cifar10_qualitative_examples} (a) we can observe that HCL assigns most importance to two pairs of negative examples from the same class ``Airplane'' and ``Truck'' (\textcolor{orange}{\textbf{orange}} bounding boxes (bboxes)) whereas \modelname selects one example from those classes and additionally selects examples from diverse classes including ``Deer'' and ``Automobile'' (\textcolor{green}{\textbf{green}} bboxes). Similar scenarios are visible in  Figure~\ref{fig:cifar10_qualitative_examples} (b), (d), (e), (f), (g), (h). Moreover, Figure~\ref{fig:cifar10_qualitative_examples} (c), (e), (f), (g) depicts that HCL assigns more importance to the negative example with the same class as the anchor, \ie ``Bird'' (Figure~\ref{fig:cifar10_qualitative_examples} (c), (e)), ``Truck'' (Figure~\ref{fig:cifar10_qualitative_examples} (f)), ``Frog'' (Figure~\ref{fig:cifar10_qualitative_examples} (g)) (\textcolor{red}{\textbf{red}} bboxes). On the other hand, \modelname avoids negative examples of the same class as the anchor by including gradient component in the importance calculation of the negative examples. However, \modelname sometimes redundant negative examples can also appear in the most five important negatives selected by \modelname because of the fewer number of ground-truth classes in \texttt{CIFAR-10} dataset (Figure~\ref{fig:cifar10_qualitative_examples} (c), (d), (h)).

\textbf{\texttt{CIFAR-100}:} In Figure~\ref{fig:cifar100_qualitative_examples} (a), (f), (h), we notice that the top five most important negative examples for HCL contain examples of the same class as the anchor \ie ``Ray'', ``Girl'', ``Possum'' (\textcolor{red}{\textbf{red}} bboxes) whereas \modelname assigns most importance to representative negative examples (\textcolor{green}{\textbf{green}}). Besides, in Figure~\ref{fig:cifar100_qualitative_examples} (b), (c), (d), (e), (g), we can see that HCL assigns most importance to redundant examples of the same image with different augmentations ((b), (c), (e), (g)) or examples of the same class ((d)) (\textcolor{orange}{\textbf{orange}} bboxes). On the other hand, \modelname avoids assigning high importance to different augmentations of the same  example or examples of the same class. 

\textbf{\texttt{TINY-IMAGENET}:} The qualitative results also follow similar trends as in \texttt{CIFAR-10} and \texttt{CIFAR-100}. For example, Figure~\ref{fig:qualitative_evaluation_tiny_imagenet_1} depicts that HCL selects redundant negative examples in the top seven most important negatives, \ie different augmentations of the same image (\textcolor{orange}{\textbf{orange}} bboxes). On the other hand, \modelname selects only one of those images (\eg class 186) and additionally selects examples from representative classes (\textcolor{green}{\textbf{green}} bboxes). Similar observations can be made in Figure~\ref{fig:qualitative_evaluation_tiny_imagenet_1} (c), Figure~\ref{fig:qualitative_evaluation_tiny_imagenet_2}. Moreover, in Figure~\ref{fig:qualitative_evaluation_tiny_imagenet_1} (b), it is noticeable that HCL assigns most importance to a negative example of the same label as the anchor (\textcolor{red}{\textbf{red}}). On the other hand, \modelname does not assign much importance to that negative example and instead selects a representative set of examples as top seven negatives.


\end{document}